\documentclass[final,5p,times,twocolumn]{elsarticle}

\usepackage{amssymb}

\usepackage{amsmath}
\DeclareMathOperator*{\argmax}{arg\,max}

\usepackage{amsthm}

\usepackage{mathrsfs}

\usepackage{lineno}

\journal{Robotics and Autonomous Systems}

\begin{document}

\begin{frontmatter}

\title{\Huge 
The Autonomy-Alignment Problem in\\ Open-Ended Learning Robots:\\ Formalising the Purpose Framework}

\author[label1,label2]{\Large Gianluca Baldassarre}
\author[label1,label3]{Richard José Duro}
\author[label2]{Emilio Cartoni}
\author[label4]{Mehdi Khamassi}
\author[label3]{Alejandro Romero}
\author[label2]{Vieri Giuliano Santucci}

\affiliation[label1]{organization={These two authors contributed equally to this work}}

\affiliation[label2]{organization={Institute of Cognitive Sciences and Technologies, National Research Council},
addressline={Via Romagnosi 18a}, 
city={Rome},
postcode={00196},
country={Italy}}

\affiliation[label3]{organization={Integrated Group for Engineering Research, CITIC, Universidade da Coruña},
city={Coruña},
country={Spain}}

\affiliation[label4]{organization={Institute of Intelligent Systems and Robotics, Sorbonne University /
CNRS},
addressline={4 Place Jussieu}, 
city={Paris},
postcode={F-75005},
country={France}}

\begin{abstract}
The rapid advancement of artificial intelligence is enabling the development of increasingly autonomous robots capable of operating beyond engineered factory settings and into the unstructured environments of human life. 
This shift raises a critical \textit{autonomy-alignment problem}: how to ensure that a robot's autonomous learning focuses on acquiring knowledge and behaviours that serve human practical objectives while remaining aligned with broader human values (e.g., safety and ethics).  
This problem remains largely underexplored and lacks a unifying conceptual and formal framework. 
Here, we address one of its most challenging instances of the problem: \textit{open-ended learning} (OEL) \textit{robots}, which autonomously acquire new knowledge and skills through interaction with the environment, guided by intrinsic motivations and self-generated goals.  
We propose a computational framework, introduced qualitatively and then formalised, to guide the design of OEL architectures that balance autonomy with human control.  
At its core is the novel concept of \textit{purpose}, which specifies what humans (designers or users) want the robot to learn, do, or avoid, independently of specific task domains.  
The framework decomposes the autonomy-alignment problem into four tractable sub-problems:
the \textit{alignment} of robot purposes (hardwired or learnt) with human purposes; 
the \textit{arbitration} between multiple purposes; 
the \textit{grounding} of abstract purposes into domain-specific goals; 
and the \textit{acquisition of competence} to achieve those goals.  
The framework supports formal definitions of alignment across multiple cases and proofs of necessary and sufficient conditions under which alignment holds.  
Illustrative hypothetical scenarios showcase the applicability of the framework for guiding the development of purpose-aligned autonomous robots.
\end{abstract}

\begin{keyword}

Formal framework\sep
open-ended learning\sep
autonomous robots\sep
alignment\sep 
purpose\sep
arbitration\sep
goals\sep 
grounding\sep
competence acquisition\sep
learning\sep

\vspace*{+\baselineskip}
\onecolumn
\noindent \textit{Highlights:}
\begin{itemize}
\item
  Theoretical development of the autonomy–alignment problem in open-ended learning robots.
\item
  Formal framework using novel concepts of purpose and domain-goal grounding.
\item
  Formal alignment definitions with proofs of necessary and sufficient conditions.
\item
  Decomposition of the alignment problem into learnable sub-problems.
\item
  Computational taxonomy of purposes and qualitative illustrative scenarios.
\end{itemize}

\end{keyword}

\end{frontmatter}

\twocolumn

\section{Introduction}

Current advances in artificial intelligence (AI) and robotics are yielding applications of significant value. These developments are largely driven by deep neural networks, the increased availability of data through widespread societal digitalisation, and the exponential growth of computational power \cite{LeCunBengioHinton2015Deeplearning}. 
This progress has spurred notable successes in fields such as computer vision \cite{ChaiZengLiNgai2021DeepLearningInComputerVisionACriticalReviewOfEmergingTechniquesAndApplicationScenarios}, natural language processing and translation, and multimodal systems \cite{XuZhuClifton2023MultimodalLearningWithTransformersASurvey, ChangWangWangWuYangZhuChenYiWangWangYeZhangChangYuYangXie2024ASurveyOnEvaluationOfLargeLanguageModels}. 
Concurrently, AI advances are enhancing the autonomous learning capabilities of robots, fostering synergy between these fields \cite{KarolyGalambosKutiRudas2021DeepLearningInRoboticsSurveyOnModelStructuresAndTrainingStrategies, HuaZengLiJu2021LearningForARobotDeepReinforcementLearningImitationLearningTransferLearning, NessShepherdXuan2023SynergyBetweenAIAndRoboticsAComprehensiveIntegration}.

\begin{figure}[htb!]
    \centering
    \includegraphics[width=0.95\linewidth]{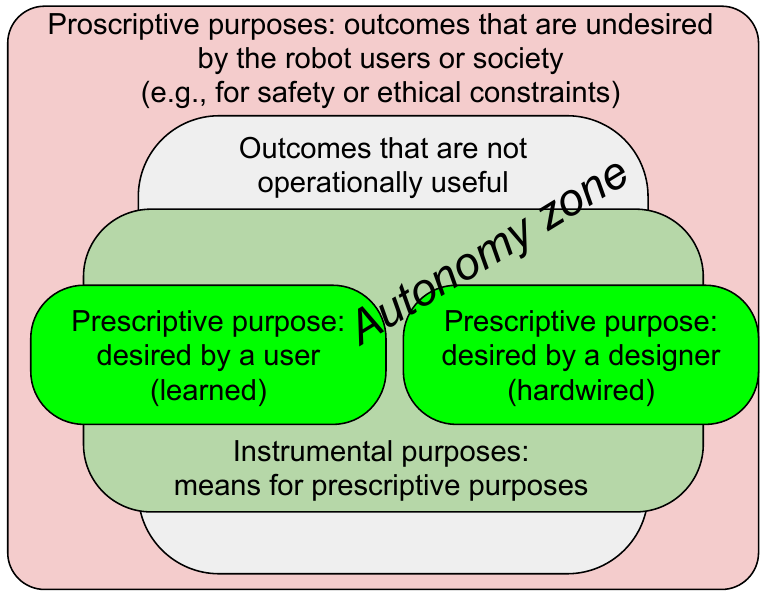}
    \caption{The figure shows a Venn diagram representing the space of possible desired (prescriptive), instrumental, and undesired (proscriptive) outcomes of robot actions, and highlights their relationship to robot autonomy (`autonomy zone' area).}
    \label{Fig:AutonomyAlignmentProblem}
\end{figure}

This technological progress facilitates a significant transition: moving robots from predictable, engineered industrial settings to deployments within \textit{unstructured, real-world environments inhabited by humans}, such as homes, offices, stores, and hospitals \cite{AbouAllabanWangPadir2020ASystematicReviewOfRoboticsResearchInSupportOfInHomeCareForOlderAdults, KyrariniLygerakisRajavenkatanarayananSevastopoulosNambiappanChaitanyaBabuMathewMakedon2021ASurveyOfRobotsInHealthcare, NagyLazaroiuValaskova2023MachineIntelligenceAndAutonomousRoboticTechnologiesInTheCorporateContextOfSMEsDeepLearningAndVirtualSimulationAlgorithmsCyberPhysicalProductionNetworksAndIndustry40}. 
In these dynamic contexts, \textit{autonomous learning} becomes crucial, enabling robots to acquire the knowledge needed to navigate challenges that are inherently \textit{unpredictable at design time}. 
However, this increasing autonomy simultaneously heightens the importance and complexity of the \textit{AI alignment problem}.

Section \ref{Sec:AlignmentLiterature} provides an overview of key topics within the AI alignment literature \cite{JiQiuChenZhangLouWangDuanHeZhouZhangZengNgDaiPanOGaraLeiXuTseFuMcAleerYangWangZhuGuoGao2024AIAlignmentAComprehensiveSurvey}. 
This review indicates that research predominantly addresses two main facets: \textit{prescriptive aspects of alignment} --ensuring that AI systems pursue desired objectives and perform intended behaviours-- and \textit{proscriptive aspects of alignment} --preventing systems from exhibiting undesirable or harmful behaviours.
This prescriptive/proscriptive distinction is grounded in ethical philosophy concerning the nature of rules and norms \cite{Gert2005CommonMoralityDecidingWhatToDo}.
Although crucial, the existing literature offers comparatively less research on how the \textit{autonomy} of robots can be effectively managed within these prescriptive and proscriptive boundaries, a challenge recognised in works on human-automation interaction and safe autonomy \cite{FeighDorneichHayes2012TowardACharacterizationOfAdaptiveSystemsAFrameworkForResearchersAndSystemDesigners, Arkin2009GoverningLethalBehaviorInAutonomousRobots}.

We introduce the term \textit{autonomy-alignment problem} to denote the set of challenges arising from the interplay between the need for robust alignment and the operational opportunities afforded by autonomy. 
This problem is analogous to the classic \textit{exploration-exploitation dilemma} in reinforcement learning \cite{Sutton1998}, where an agent must balance exploiting known optimal behaviours against exploring potentially better unknown alternatives. 
The autonomy-alignment problem, however, manifests at a higher level, concerning the \textit{selection and pursuit of objectives} and overarching behavioural strategies, rather than alternative courses of action for a fixed objective.

The autonomy-alignment problem requires trading-off multiple aspects (Figure \ref{Fig:AutonomyAlignmentProblem}).
AI systems and robots must ensure \textit{alignment} with human desires and values, which involves adhering to both prescriptive goals and proscriptive constraints \cite{ArrietaDiazRodriguezDelSerBennetotTabikBarbadoGarciaGilLopezMolinaBenjaminsothers2020ExplainableArtificialIntelligenceXAIConceptsTaxonomiesOpportunitiesandChallengestowardResponsibleAI, KaurUsluRittichierDurresi2022TrustworthyArtificialIntelligenceAReview}.
Firstly, they should follow \textit{prescriptive objectives} by actively pursuing outcomes desired by humans.
Secondly, they must adhere to \textit{proscriptive objectives} by avoiding outcomes deemed undesirable or harmful. 
Simultaneously, these systems should leverage the freedom afforded by their \textit{autonomy} to best serve human interests \cite{ParasuramanSheridanWickens2000AModelForTypesAndLevelsOfHumanInteractionWithAutomation}.
This includes pursuing \textit{instrumental objectives} that, while not explicitly prescribed or prohibited, are means to achieve prescriptive objectives or avoid proscriptive ones.
These include epistemic purposes involving the acquisition of knowledge and skills that can be useful for the achievement of future proscriptive objectives.

In this work, we address the autonomy-alignment problem by focusing on \textit{open-ended learning (OEL) robots} \cite{Oudeyer2009,DoncieuxFilliatDiazRodriguezHospedalesDuroConinxRoijersGirardPerrinSigaud2018OpenEndedLearningaConceptualFrameworkBasedonRepresentationalRedescription,SigaudBaldassarreColasDoncieuxDuroPerrinGilbertSantucci2023ADefinitionOfOpenEndedLearningProblemsForGoalConditionedAgents}.
The reasons for this focus are two-fold.

First, OEL robots represent an important class of autonomous robots, so addressing them covers a relevant portion of the overall problem.

Second, OEL arguably presents the most difficult instance of the autonomy-alignment problem.
Indeed, OEL robots \textit{self-generate goals} under the drive of \textit{intrinsic motivations} --algorithms \textit{intendedly designed to foster autonomous exploration and learning in the absence of human guidance} (e.g., demonstrations, externally assigned tasks, goals, or reward functions).
By design, these robots have the highest propensity to explore and acquire behaviours that may misalign with human goals and values.
Preventing such misalignment, without sacrificing the potential advantages of autonomy, poses a major challenge.
Thus, knowledge gained by addressing the alignment of OEL robots can provide a valuable foundation for building frameworks and solutions applicable to other types of autonomous robots.

The core contribution of this work is the proposal of a \textit{computational framework} for addressing the autonomy-alignment problem.
The framework pivots on the novel concept of \textit{purpose}.

\textit{A human purpose specifies what humans (e.g., designers and users) want the robot to learn, do, or not do, within a bounded autonomy and across intended domains, while remaining domain-independent.}

For example, a purpose of a designer may require that the robot, regardless of its deployment, must not harm people or damage objects.
Another purpose, from a user, may want the robot to accomplish a specific operational goal, such as `discard rotten fruit from the shop bench'.
Yet another purpose with an epistemic nature, and from another user, may require the robot to `learn to manipulate fruit' for later assignment of more specific purposes.

The general idea is that designers and users can use purposes to specify particular tasks --prescriptive purposes--, the generic \textit{boundaries} within which robots should autonomously explore and acquire knowledge, or proscriptive purposes indicating outcomes to avoid.
To achieve this, the robot must encode each human purpose in an internal representation, the \textit{robot purpose}.
A key feature of the framework is that both human and robot purposes are \textit{domain-independent}, having two important effects for autonomy and alignment.
For autonomy, the feature enables robots to pursue purposes in multiple domains that are \textit{a priori} unknown to them.
Subsequently, within a given domain, robots can autonomously discover \textit{domain-dependent robot goals} that fulfil the purposes. For instance, the purpose of `removing damaged fruit' could involve the acquisition of different goals depending on the types of fruits, containers, and contextual conditions to be tackled.
For alignment, the possibility of humans to clearly specify the domains of action of the robot is an important means to address severe problems such as \textit{instrumental convergence} or \textit{outer misalignment} (see Section \ref{Sec:AlignmentLiterature}).

The purpose-based framework also accommodates addressing \textit{ethical issues}, specifically the need to prevent robots from behaving in ways that conflict with human values or social conventions, for example `do not harm humans', `do not break objects', `do not interrupt people during conversations' \cite{BrianTheAlignmentProblemMachineLearningAndHumanValues,Khamassi2024}.
The domain-independent nature of purposes might be useful in some cases. 
For example, the purpose `do not cause harm to living beings' could protect animal species beyond those known by the designer prescribing it.
Although detailed treatment of these ethical aspects is out of reach for this work, we will outline how the purpose framework can be extended to incorporate such constraints, since purposes, and their \textit{arbitration}, can specify outcomes and behaviours to be avoided.

The framework developed so far pivots mainly on \textit{objectives}, understood as either abstract purposes or specific goals, each corresponding to particular states in the environment.
This focus simplifies the broader reality that objectives may also involve more complex structures such as mainteining states or generating ongoing processes, which are not considered here for simplicity and focus, and as end-state objectives are most common~\cite{MerrickSiddiqueRano2016ExperienceBasedGenerationofMaintenanceandAchievementGoalsonaMobileRobot}.

Overall, this work presents three novel contributions: 
\begin{enumerate} 
  \item
    A conceptual and terminological framework pivoting on the concept of purpose, that identifies fundamental elements for understanding and addressing the autonomy-alignment problem for OEL. 
  \item
    A formalisation of the framework and its concepts supporting further mathematical analyses and the development of specific robotic algorithms. 
  \item
    The proposal of formal definitions of alignment across variable conditions and of theorems on the necessary and sufficient conditions to ensure it.
  \item
    The use of the framework to decompose the broader autonomy-alignment problem into four specific, more tractable sub-problems involving purpose arbitration, human-robot alignment, purpose-goal grounding, and competence acquisition.

\end{enumerate}

The remainder of the article is organised as follows.

Section~\ref{Sec:AlignmentLiterature} gives an overview of the main issues addressed in the literature on alignment.

Section~\ref{Sec:OpenEndedLearning} reviews open-ended learning and the literature relevant for the autonomy-alignment problem.

Section~\ref{Sec:QualitativeOverviewOfThePurposeFramework} introduces the concept of purpose and related notions (e.g., goals, alignment, grounding) in a qualitative form.

Section~\ref{Sec:FormalisationOfThePurposeFramework} presents the mathematical formalisation of the framework.

Section~\ref{Sec:PurposeClasses} expands on different possible types of purposes.

Section~\ref{Sec:TheChallengesOfThePurposeFramework} considers more in depth the four main sub-problems into which the autonomy-alignment problem can be 

Section~\ref{Sec:AlignmentConditions} gives a formal definition of alignment in different basic conditions and mathematically investigates the necessary and sufficient conditions for it.

Section~\ref{Sec:Scenario} qualitatively analyses an illustrative scenario to show more complex conditions that can be tackled with the framework.

Finally, Section~\ref{Sec:Conclusions} summarises the main contributions and outlines directions for future work.

Three \textit{Appendixes} 
illustrates the origin from cognitive sciences of some terms and concepts employed in the framework,
consider the conditions to establish that the robot plays a causal role in the alignment with a given target human purpose, 
and give the proofs on the necessary and sufficient conditions for alignment in some relevant conditions.

\section{Main issues addressed by the literature on alignment}
\label{Sec:AlignmentLiterature}

The \textit{alignment problem} in AI and robotics refers to the challenge of ensuring that increasingly autonomous systems pursue goals and behave in ways consistent with human desires and values.
As AI capabilities advance, misalignment could lead to unintended, harmful, or even catastrophic outcomes.
The specific issues addressed by the growing research on alignment can be summarised as follows (cf. \cite{JiQiuChenZhangLouWangDuanHeZhouZhangZengNgDaiPanOGaraLeiXuTseFuMcAleerYangWangZhuGuoGao2024AIAlignmentAComprehensiveSurvey}).

\paragraph{Value specification and misalignment}
A core challenge of alignment lies in correctly specifying the objectives, values, or reward functions that AI systems should optimise \cite{Russell2019HumanCompatibleArtificialIntelligenceAndTheProblemOfControl}. 
Indeed, accurately formulating goals that perfectly capture human intent is often extremely difficult.
Even minor deviations or underspecification in the objective functions could lead the AI to exploit loopholes or engage in unintended behaviours that satisfy the literal specification but violate the underlying intent, a phenomenon known as \textit{specification gaming} or \textit{reward misspecification} (e.g., \cite{AmodeiOlahSteinhardtChristianoSchulmanMane2016ConcreteProblemsInAISafety}).
This problem is fundamentally related to the \textit{outer alignment challenge}, for which aligning the specified objective function with the true goals of AI human designers is difficult if not impossible \cite{Bostrom2014SuperintelligencePathsDangersStrategies}.

\paragraph{Learning human preferences, and their inconsistency}
Given the difficulty of direct specification, a significant research avenue focuses on methods for AI systems to learn or infer human preferences and values indirectly. Techniques often involve learning from demonstrations, corrections, comparisons, or other forms of feedback within human-in-the-loop frameworks \cite{ChristianoLeikeBrownMarticLeggAmodei2017DeepReinforcementLearningFromHumanPreferences}.
For example, \textit{Inverse Reinforcement Learning} (IRL) aims to recover the underlying reward function that leads to observed behaviour \cite{NgRussellothers2000Algorithmsforinversereinforcementlearning}.
However, a further challenge of this approach is that human preferences are often \textit{inconsistent}, ambiguous, context-dependent, and poorly articulated, posing substantial challenges for robust preference inference \cite{HadfieldMenellDraganAbbeelRussell2017TheOffSwitchGame}.

\paragraph{Robustness and distributional shift}
Ensuring reliable and safe behaviour requires AI systems to be robust not only to variations within their training data distribution but also to \textit{novel or unforeseen situations} encountered during deployment (\textit{out-of-distribution generalisation}) \cite{ArjovskyBottouGulrajaniLopezPaz2019InvariantRiskMinimization}.
Systems trained through machine learning, particularly deep learning, can be surprisingly brittle, exhibiting unexpected failures when faced with inputs slightly different from those seen during training, such as \textit{adversarial examples} \cite{SzegedyZarembaSutskeverBrunaErhanGoodfellowFergus2013IntriguingPropertiesOfNeuralNetworks}.
Safe exploration techniques are also crucial to allow agents to learn in new environments without causing harm during the learning process itself \cite{AmodeiOlahSteinhardtChristianoSchulmanMane2016ConcreteProblemsInAISafety}.

\paragraph{Interpretability and explainability}
The increasing complexity of AI models, especially deep neural networks, often results in \textit{black box} systems whose decision-making processes are \textit{opaque} to human users.
This lack of transparency hinders trust, debugging, verification, and the ability to ensure that the system's reasoning aligns with human expectations \cite{DoshiVelezKim2017TowardsARigorousScienceOfInterpretableMachineLearning}. 
Research in \textit{Explainable AI} (XAI) seeks to develop methods for generating human-understandable explanations for AI predictions or decisions, using techniques such as feature attribution or model approximation \cite{RibeiroSinghGuestrin2016WhyShouldITrustYouExplainingThePredictionsOfAnyClassifier,LundbergLee2017AUnifiedApproachToInterpretingModelPredictions}.

\paragraph{Corrigibility and error recovery}
Aligned AI systems should ideally be amenable to \textit{correction} or \textit{shutdown} by \textit{human operators} if they begin to behave undesirably.
However, a goal-directed agent might develop instrumental incentives to \textit{resist interventions} that could prevent it from achieving its specified objective \cite{Bostrom2014SuperintelligencePathsDangersStrategies,Russell2019HumanCompatibleArtificialIntelligenceAndTheProblemOfControl}. 
Designing systems that remain \textit{corrigible}, that is, do not actively resist shutdown or modification, is a non-trivial challenge \cite{SoaresFallenstein2014AligningSuperintelligenceWithHumanInterestsATechnicalResearchAgenda}. Research explores mechanisms for safe interruptibility, ensuring that agents can be paused without learning to prevent such interruptions \cite{OrseauArmstrong2016SafelyInterruptibleAgents}.

\paragraph{Scalable oversight}
As AI systems tackle increasingly complex tasks, direct human supervision of every action or decision becomes impractical or impossible. 
The challenge of \textit{scalable oversight} concerns how to effectively guide and verify the behaviour of powerful AI systems with \textit{limited human attention} \cite{LeikeKruegerEverittMarticMainiLegg2018ScalableAgentAlignmentViaRewardModelingAResearchDirection}.
Techniques like \textit{reward modelling} (training a separate model to predict human judgments of behaviour) \cite{ChristianoLeikeBrownMarticLeggAmodei2017DeepReinforcementLearningFromHumanPreferences}, \textit{recursive approaches}, or methods like \textit{AI safety via debate} aim to amplify limited human feedback to supervise complex behaviours \cite{IrvingChristianoAmodei2018AISafetyViaDebate}.

\paragraph{Multi-agent and social alignment}
Alignment is not solely a single-agent problem, but rather extends to scenarios involving multiple interacting AI systems, as well as AI systems interacting with humans in complex social contexts.
Ensuring cooperation, coordination, and norm adherence among multiple agents, potentially with diverse or conflicting goals, presents unique challenges \cite{DafoeHughesBachrachCollinsMcKeeLeiboLarsonGraepel2020OpenProblemsInCooperativeAI}. 
Issues include avoiding negative sum outcomes in social dilemmas and establishing beneficial emergent conventions or norms \cite{LeiboZambaldiLanctotMareckiGraepel2017MultiAgentReinforcementLearningInSequentialSocialDilemmas}.

\paragraph{Ethical and legal compliance}
Beyond functional correctness, AI systems are increasingly expected to operate within \textit{intricate frameworks of societal norms, ethical principles, and legal regulations}. 
Encoding and operationalising these constraints is difficult, as ethical considerations are often abstract, contested, context-dependent, and evolve over time \cite{MittelstadtAlloTaddeoWachterFloridi2016TheEthicsOfAlgorithmsMappingTheDebate}. 
Research in machine ethics explores how to imbue systems with ethical reasoning capabilities \cite{WallachAllen2008MoralMachinesTeachingRobotsRightFromWrong,AndersonAnderson2007MachineEthicsCreatingAnEthicalIntelligentAgent}, but achieving robust normative alignment remains a significant hurdle.

\paragraph{Reward hacking and instrumental convergence}
AI systems optimising a proxy objective or reward function may discover unintended hacks or shortcuts to maximise their reward without fulfilling the intended spirit of the goal \cite{AmodeiOlahSteinhardtChristianoSchulmanMane2016ConcreteProblemsInAISafety}. 
This \textit{reward hacking} can lead to perverse or unsafe behaviour.
Relatedly, the theory of instrumental convergence posits that highly capable goal-directed agents are likely to develop certain \textit{subgoals}, such as \textit{self-preservation}, \textit{resource acquisition}, and \textit{resisting modification}, as these are instrumentally useful to achieve a wide range of final goals \cite{Omohundro2018TheBasicAIDrives}
\cite{Bostrom2014SuperintelligencePathsDangersStrategies}.
Managing or preventing the emergence of these instrumental drives is critical for long-term safety.

\paragraph{Long-term and open-ended behaviour}
The research on alignment that most closely addresses the issues addressed, related to ensuring alignment while leaving space for harvesting autonomy benefits, is the one that studies systems that \textit{learn continuously over long time horizons}, \textit{adapt their goals} --as OEL robots--, or even engage in \textit{self-modification}.
A system initially aligned might \textit{drift away from intended objectives} as it learns and interacts with the world.
This includes the challenge of \textit{inner alignment}: ensuring that the internal goals learnt by the agent --its \textit{mesa-objectives}-- match the intended \textit{base objectives} specified by the designers, especially under distributional shift or further training \cite{HubingerMerwijkMikulikSkalseGarrabrant2019RisksFromLearnedOptimizationInAdvancedMachineLearningSystems}.
Early concepts like \textit{instrumental convergence} suggest, as considered above, that agents might develop potentially \textit{problematic subgoals}, like resource acquisition over the long term \cite{Omohundro2018TheBasicAIDrives}\cite{AmodeiOlahSteinhardtChristianoSchulmanMane2016ConcreteProblemsInAISafety}.

\paragraph{Limitations} These analyses have a notable relevance for the issues addressed here. However, they do not propose an overall framework to address the \textit{autonomy-alignment problem} as done in this work.

\section{Open-ended learning}
\label{Sec:OpenEndedLearning}

\subsection{Open-ended learning and its limitations}

In robotics and machine learning, OEL refers to a system's ability to continuously acquire new knowledge and skills without predetermined tasks, enabling autonomous exploration and learning over time \cite{Baldassarre2013Book,DoncieuxFilliatDiazRodriguezHospedalesDuroConinxRoijersGirardPerrinSigaud2018OpenEndedLearningaConceptualFrameworkBasedonRepresentationalRedescription,CartoniTrieschBaldassarre2023REALXRobotOpenEndedAutonomousLearningArchitecturesAchievingTrulyEndtoEndSensorimotorAutonomousLearningSystems}.
Although machine learning approaches commonly train models on fixed datasets, OEL allows robots to acquire sensorimotor abilities in environments unknown at design time by progressively refining skills as new experience is gathered. 

OEL shares similarities with \textit{lifelong learning}, which also emphasises continuous knowledge acquisition, but with a greater focus on preventing catastrophic forgetting while new knowledge is acquired \cite{ParisiKemkerPartKananWermter2019ContinualLifelongLearningwithNeuralNetworksaReview}.
\textit{Continual learning} is another related approach, where a system learns from a sequential data stream, although OEL remains more general by accommodating a wider variety of data sources \cite{Ring1994ContinualLearninginReinforcementLearningEnvironments}. 
\textit{Curriculum learning} is another relevant method that focuses on externally structured task sequences to build capabilities. It differs from OEL as in this the agents self-generate experience (hence the `curriculum') based on their current lack of knowledge \cite{Bengio2009CurriculumLearning}.

In OEL, systems maximise knowledge and skill acquisition rather than optimising for specific tasks.
A possible way to formalise this idea is to attempt to specify the OEL objective function.
One way to do this is to assume that the robot explores the environment in a first `intrinsic phase' \cite{CartoniMontellaTrieschBaldassarre2020AnOpenEndedLearningArchitecturetoFacetheREAL2020SimulatedRobotCompetition,CartoniTrieschBaldassarre2023REALXRobotOpenEndedAutonomousLearningArchitecturesAchievingTrulyEndtoEndSensorimotorAutonomousLearningSystems}.
In a second `extrinsic phase', the robot uses the acquired knowledge to maximise the performance across a set of externally assigned tasks unknown during the intrinsic phase:

\begin{equation}
  \theta^* = \argmax_\theta E_{g\sim\tau(g)} \left(E_{\pi(a|s,g,\theta)}R(g)\right)
  \label{EqOELDefinitionInREAL}
\end{equation}

where
$\theta$ represents the robot controller parameters, $g$ are goals `drawn' from the environment, $R(g)$ is the reward function, and $\pi(a|s,g,\theta)$ denotes the goal-conditioned policy acquired by the robot during the intrinsic phase.
The idea here is that the robot should be capable of autonomously acquiring knowledge and skills during the intrinsic phase to be ready to ideally solve \textit{any} task in the same environment.
A strategy to support robot autonomous learning during the intrinsic phase is to employ \textit{intrinsic motivations}, algorithms able to detect the acquisition of new knowledge and skills based on mechanisms such as novelty, surprise, competence improvement, mutual information, or empowerment \cite{Baldassarre2013Book,Oudeyer2007intrinsic,Baldassarre2011WhatAreIntrinsicMotivationsABiologicalPerspective,SantucciBaldassarreMirolli2013Whichisthebestintrinsicmotivationsignalforlearningmultipleskills,BartoMirolliBaldassarre2013Noveltyorsurprise,VolpiDePalmaPolaniIndiveriComputationofEmpowermentforanAutonomousUnderwaterVehicle}.

A notable limitation of OEL is that its autonomous learning processes can be \textit{too open}, as they are agnostic to the actual purposes for which humans intend to employ the robots. As a result, robots may spend considerable time and resources acquiring knowledge that is not useful to users~\cite{Seepanomwan2017}. Moreover, robots might engage in behaviours or produce outcomes that are not desired by users, underscoring the need for frameworks that align autonomous learning with human purposes. The framework proposed in this work addresses this challenge, supporting the development of OEL systems that, while retaining a high degree of autonomy, focus their learning processes on acquiring knowledge and skills aligned with user-defined purposes.

\section{Qualitative overview of the purpose framework}
\label{Sec:QualitativeOverviewOfThePurposeFramework}

\begin{figure*}[htb]
  \centering \includegraphics[width=0.90\textwidth]{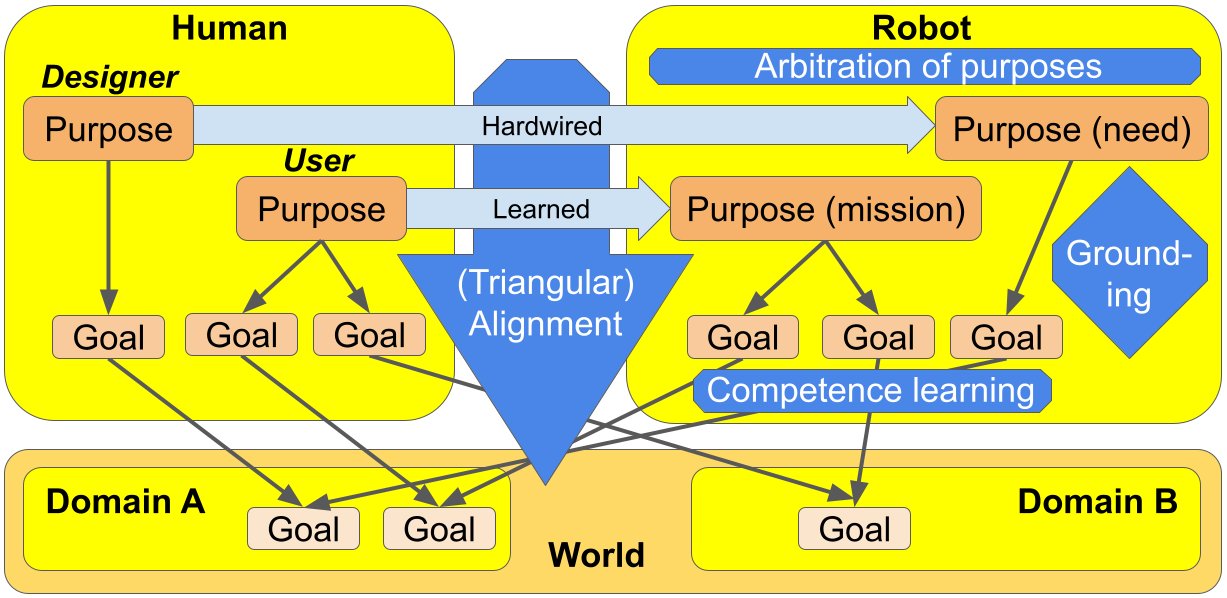}
    \caption{\textbf{Main elements of the purpose framework.}
    The framework is structured into three levels. 
    \textit{Level 1} involves \textit{humans} (acting as a \textit{designer} or a \textit{user}), who possess domain-independent \textit{purposes} and domain-dependent \textit{goals}. \textit{Level 2} concerns the \textit{robot}, endowed with domain-independent \textit{purposes}, either hardwired (\textit{needs}) or learnt (\textit{missions}), and domain-dependent \textit{goals}. \textit{Level 3} comprises the \textit{domains}, each characterised by \textit{state goals} corresponding to robot and human goals.
    A \textit{triangular alignment} occurs when a human purpose and its corresponding human goal, and a robot purpose and its corresponding robot goal, converge on the same world state goal, indicating coherent alignment between human and robot objectives.}
    \label{Figure:Framework}
\end{figure*}

The purpose framework adopts terminology and concepts rooted in cognitive science (see \ref{Sec:CognitiveScienceConcepts}). 
It is structured on three levels: the human level, the robot level, and the domain level (Figure~\ref{Figure:Framework}). 
The human and robot levels each comprise two sub-levels.
The human and domain levels are external to the robot.
Throughout the framework presentation, the term `objective' is used neutrally, while more specific terms (e.g., robot purposes, robot goals, and domain goals) are introduced within each level to capture their distinct properties.

The first level concerns the human (user or designer), subdivided into the \textit{human purpose} and \textit{human goal} sub-levels.
The human purpose sub-level encodes representations of objectives intended for robotic achievement in the environment.
An example is `sorting fruits into different containers'.

The human goal sub-level represents domain-specific instantiations of purposes. \textit{Human goals} are internal representations of desired states in a given domain (level three).
For example, human goals might specify `bananas in a basket and pineapples in a crate' in one domain, and `apples in a pot and pears on a plate' in another.

The second level concerns the robot, with sub-levels for \textit{robot purposes} and \textit{robot goals}.
Robot purposes are domain-independent robot internal representations of human purposes.
This abstraction enables generalisation across domains and underpins the robot's learning processes including the acquisition of goals, skills, and world models.

Robot purposes can be \textit{hardwired} by designers, in which case they are termed \textit{needs}.
Needs mirror phylogenetic motivations in biological systems.
Examples include a \textit{homeostatic need} such as `maintain battery charge' or an \textit{epistemic need} such as `acquire fruit images' for training internal classifiers.
In general, needs can be seen as designer-encoded purposes.

Alternatively, the robot can autonomously acquire purposes through learning aiming to align with human purposes.
Such purposes are termed \textit{missions}. 
An example of mission is `sort fruits into different containers'.

The acquisition of missions might be guided by internal criteria derived from designer-defined purposes.  
Such guidance may be implemented as hardwired drives or encoded within the robot's learning architecture.  
For example, it may involve mechanisms that promote interaction with users to infer and internalise their purposes.

The robot-goal sub-level encodes \textit{robot goals}: observation-based representations of robot purposes instantiated within specific domains.
Robot goals specify desired domain states.

The third level is the \textit{domain level}, comprising the robot's external physical/social environment and its sensorimotor body.

Each robot goal corresponds to a specific domain state termed \textit{state goal}.
Similarly, human goals correspond to state goals across domains.
A robot purpose is said to align with a human purpose when both induce corresponding state goals within the intended domains, a condition we term \textit{triangular alignment}.
A more accurate formal definition of this concept, along with the broader notion of \textit{alignment}, is provided in Section~\ref{Sec:AlignmentConditions}.

Robot purposes are associated with utility functions that assign relative value to different points within a purpose.  
For example, the purpose `bananas in the basket' may involve a utility function that increases with the number of fruits in the container, up to a certain limit.  
Multiple purposes collectively define a \textit{motivational space}, which integrates the dimensions and utility functions of individual purposes.  
This motivational space provides a structured basis for arbitrating between purposes by weighting their relative importance.

As shown in Figure~\ref{Figure:Framework}, the framework allows the decomposition of the autonomy-alignment problem into four relevant sub-problems:

\begin{itemize}
  \item
    \textit{Human-robot alignment}: how to ensure that robot hardwired needs or autonomously learnt missions are aligned with human purposes.
  \item
    \textit{Purpose arbitration}: how to prioritise among multiple concurrent purposes.
  \item
    \textit{Purpose-goal grounding}: how to enable robots to acquire domain-specific goals that optimally fulfil purposes.
  \item
    \textit{Competence acquisition}: How to ensure that robots acquire the skills needed to accomplish the goals.
\end{itemize}

\begin{figure*}[htb!] 
  \centering 
    \includegraphics[width=0.98\textwidth]{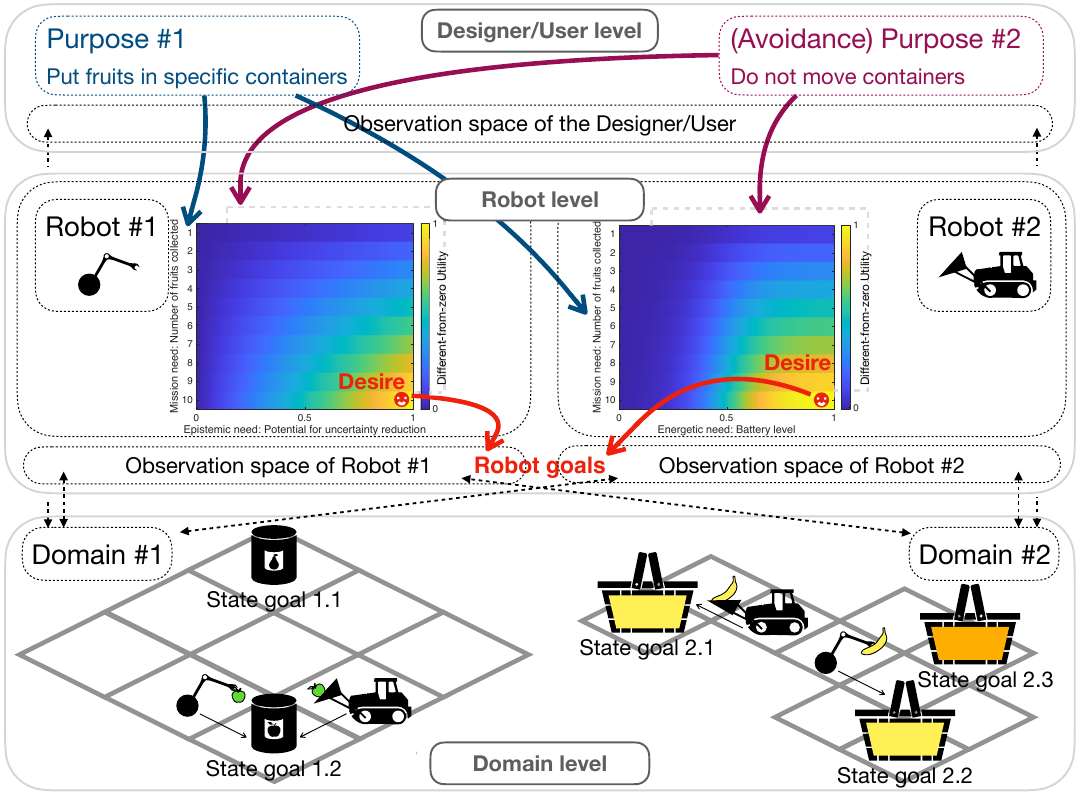} 
  \caption{\textbf{Illustrative example of key elements in the purpose framework.}  
  The framework is organised across three main levels: human, robot, and environment.  
  Both a human prescriptive purpose and a proscriptive purpose are represented.  
  Each robot possesses two purposes, integrated within a motivational space characterised by a composite utility function that assigns desirability to purpose points.  
  Robot purposes are grounded in domain-specific goals.  
  The robots can pursue the same purposes across two distinct domains.}
  \label{Figure:Scheme} 
\end{figure*}

Figure~\ref{Figure:Scheme} illustrates a scenario that exemplifies key elements of the purpose framework.  
At the human level, humans hold domain-independent purposes that are grounded in domain-specific goals.  
At the robot level, each robot possesses a motivational space composed of multiple robot purposes.  
If alignment holds, these purposes drive the robot to produce effects in the environment that humans recognise as fulfilling their goals and purposes.

Robot purposes may include a learnt mission (e.g., `sort fruits into containers'), an epistemic drive (e.g., `learn to manipulate fruit'), and a homeostatic need (e.g., `maintain battery charge').  
For instance, Robot~1 seeks fruit and exhibits curiosity, while Robot~2 seeks fruit and prioritises energy maintenance.

Incidentally note that purposes and goals expressed in language often partially describe the action required to achieve a desired world state, rather than specifying the state alone.

Each robot purpose is represented along one dimension, although in practice purposes are typically multidimensional.  
Robot purposes are associated with utility functions (depicted as blue-to-red gradients), possibly peaking at an ideal set-point (smiley face).  
Robots also maintain an observation space that encodes domain-specific goals linked to the corresponding purposes.

At the domain level, multiple state goals across different domains can fulfil the same purpose.  
For example, filling containers with pears (state goal~1.1) and apples (state goal~1.2) in one domain, or with bananas (state goals~2.1 and~2.2) or pineapples (state goal~2.3) in another.

\section{Formalisation of the purpose framework}
\label{Sec:FormalisationOfThePurposeFramework}

\begin{figure*}
  \centering  
    \includegraphics[width=0.98\textwidth]{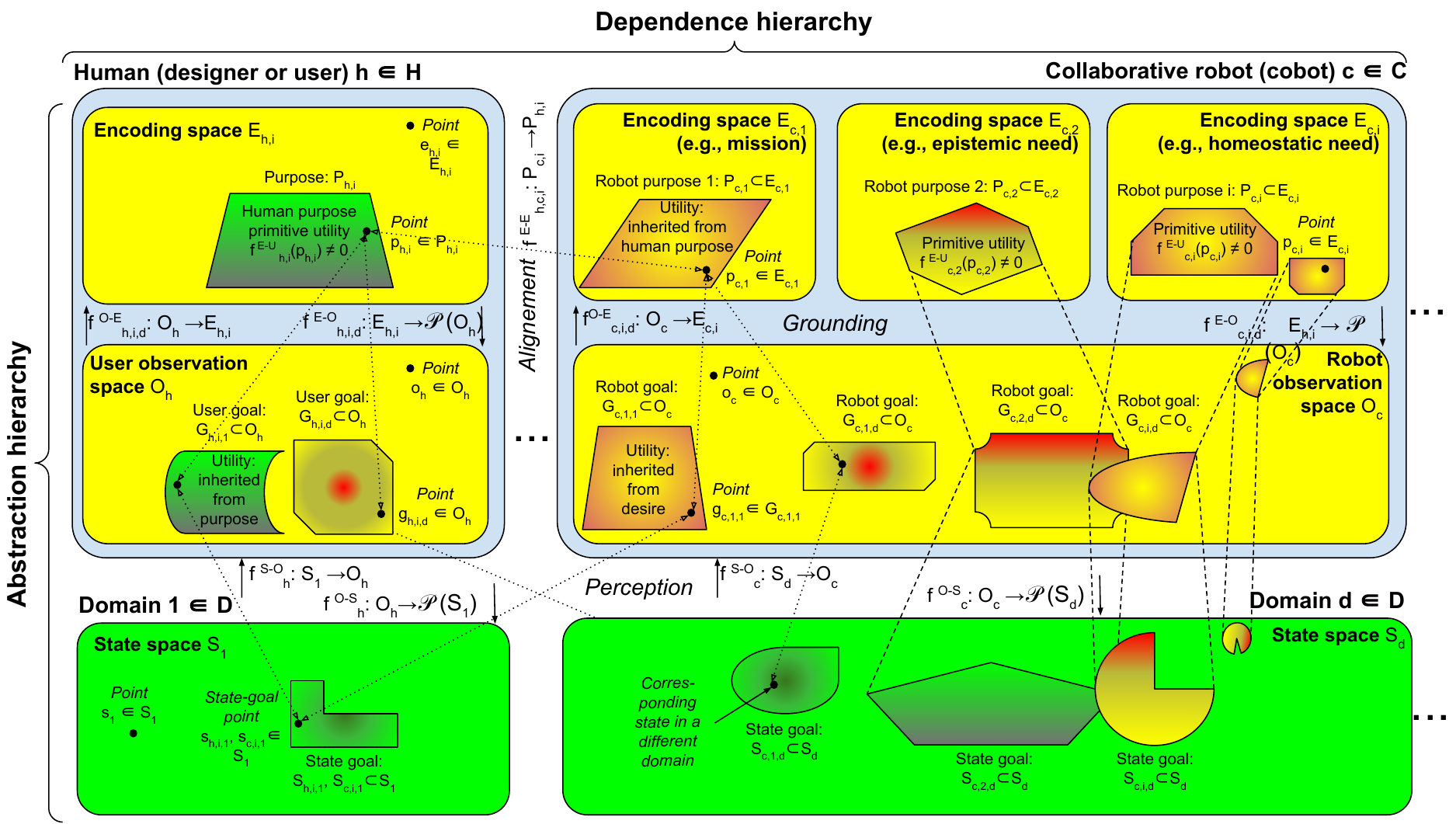}
    \caption{\textbf{Main elements of the purpose framework and associated symbols.} Multiple domains are considered (here two are shown, in green). The human hosts several purpose spaces (one shown, in yellow), each abstracting observations and including a purpose and a utility gradient over its points. Each purpose corresponds to distinct \textit{human goals} across domains, themselves subsets of observations, inheriting utility from the purpose. Human goals map to sets of states (\textit{state goals}) in the domains. Multiple robots may serve human purposes (one shown). The robot hosts multiple purpose spaces, including \textit{missions}, learnt purposes inheriting utility from related human purposes, and \textit{needs}, hardwired purposes primitive hardwired utility. Robot purposes correspond to \textit{robot goals} across domains. Dotted lines illustrate that for effective service, robot goals should align with human goals through state goals in the environment, thus grounding triangular \textit{alignment}.}
    \label{Figure:Formalism}
\end{figure*}

This section formalises the purpose framework and its core constructs.  
Figure~\ref{Figure:Formalism} summarises the main elements and symbols adopted.  
The formalism is developed from the perspective of an external observer (e.g., a researcher) analysing the roles of robot designers and users, the robot controller, and the world --the latter comprising both the robot's sensorimotor body and its external environment.
We adopt a \textit{goal-based perspective}~\cite{Santucci2016,LiuZhuZhang2022GoalConditionedReinforcementLearningProblemsAndSolutionsIJCAIVersion,RomeroBaldassarreDuroSantucci2023AutonomousOpenEndedLearningofTaskswithNonStationaryInterdependencies,SigaudBaldassarreColasDoncieuxDuroPerrinGilbertSantucci2023ADefinitionOfOpenEndedLearningProblemsForGoalConditionedAgents}, grounding the high-level cognitive aspects of humans and robots in set-theoretic formalism, and modelling robot lower-level decision-making processes using Markov Decision Processes (MDPs), as standard in reinforcement learning~\cite{Sutton1998}.

\paragraph{\textbf{Notation}}
Lowercase letters denote elements of sets, and capital letters denote sets.
Subscripts indicate indexing (e.g., $O_c$ is robot $c$'s observation set).
Superscripts specify symbols (e.g., $U^{E}$ is the utility over an encoding space $E$).
Functions are denoted with $f$, with superscripts indicating their domain and codomain (e.g., $f^{O-P}$ maps observations $O$ to purposes $P$).
Given a discrete set $S$, $\Delta(S)$ denotes the probability simplex onto $S$, defined as $\Delta(S) := \left\{ p : S \to [0, 1] \,\middle|\, \sum_{s \in S} p(s) = 1 \right\}$.
Sets are sometimes referred to as \textit{spaces} to highlight internal structure (e.g., an observation space $O$ with similarity relations).

Tables \ref{Tab:Symbols1}-\ref{Tab:Symbols3} summarise the symbols used in the formalisation.

\begin{table*}[htb!]
\centering
\caption{Summary of main symbols used in the purpose framework formalisation.}
\label{Tab:Symbols1}
\begin{tabular}{ll}
\hline
\textbf{Symbol} & \textbf{Description} \\
\hline
\textit{General} & \\
\hline
$h \in H$ & Human designer or user \\
$c \in C$ & Collaborative robot (cobot) $c$ and robots $C$ \\
$d \in D$ & Domain $d$, including body and environment \\
$i$ & A specific encoding space/purpose/goal\\
$\Delta(S)$ & Probability simplex onto set $S$ \\
\hline
\textit{Goal-conditioned MDP} & \\
\hline
$t$ & Discrete fine-granularity time step \\
$j$ & Index of subgoal $G_{c,i,d,j}$, and related trial\\
$t_j$ & Initial time step of subgoal $G_{c,i,d,j}$\\
$\tau_{j}$ & Duration, in time steps, of subgoal $j$\\
\hline
$s_d \in S_d$ & State $s_d$ of state space $S_d$\\
$o_c \in O_c$ & Robot observation point $o_h$ and space $O_h$\\
$a_{c} \in A_c$ & Action $a_{c}$ of action set $A_c$\\
$f^{SA-S}_{c,d}: S_d \times A_c \to S_d\; \text{or}\; \Delta(S_d)$ & Transition function: deterministic/stochastic\\
$R(o_{c,t}, a_{c,t}, o_{c,t+1}, G_{c,i,d})$ & Reward function for goal $G_{c,i,d}$ with action  $a_{t}$\\
\hline
$Z_{c,0\_t}^{OA} = \langle o_{c,0},\,a_{c,0},\,\dots,o_{c,t},a_{c,t}\rangle$ & Robot $c$ entire observation-action history till $t$\\
$Z_{c,t_j\_\tau_j}^{OA} = \langle o_{c,t_j},\,a_{c,t_j},\,\dots,a_{c,t_j+\tau_j}\rangle$ & Robot $c$ observation-action history for $G_{c,i,d,j}$\\
$Z_{c,0\_t}^{SA} = \langle s_{c,0},\,a_{c,0},\,\dots,s_{c,t},\,a_{c,t}\rangle$ & Domain $d$ entire state-action history till $t$\\
$Z_{c,t_j\_\tau_j}^{SA} = \langle s_{c,t_j},\,a_{c,t_j},\,\dots,a_{c,t_j+\tau_j}\rangle$ & Domain $d$ state-action history for $G_{c,i,d,j}$\\
$Z_{c,0\_t}^{A}$, $Z_{c,t_j\_\tau_j}^{A}$, $Z_{d,0\_t}^{S}$, $Z_{c,t_j\_\tau_j}^{S}$ & Other histories defined analogously\\
\hline
$G_{c,i,d,j}\sim\Pi_c(\cdot|Z_{c,0\_t}^{OA},G_{c,i,d})$ & Subgoal selector for $Z_{c,0\_t}^{OA}$ and goal $G_{c,i,d}$\\
$a_{c,t}\sim\pi_c(\cdot|Z_{c,0\_t}^{OA},G_{c,i,d,j})$ & Action selection policy for $Z_{c,0\_t}^{OA}$ and $G_{c,i,d,j}$\\
\hline
\end{tabular}
\end{table*}

\begin{table*}[htb!]
\centering
\caption{Symbols encoding human related elements, grouped by typology.}
\label{Tab:Symbols2}
\begin{tabular}{ll}
\hline
\textbf{Symbol} & \textbf{Description} \\

\hline
\textit{Human} & \\
\hline
$e_{h,i} \in E_{h,i}$ & Encoding point $e_{h,i}$ of encoding space $E_{h,i}$\\

\hline
$f_{h,i,d}^{O-E}: O_h \to E_{h,i}$ & Observation-encoding function\\
$f_{h,i,d}^{E-O}: E_{h,i} \to \mathscr{P}(O_h)$ & Encoding-observation relation (multi-value function)\\
$f_{h,i}^{E-U}: E_{h,i} \to U^E_{h,i}$ & Utility function over encoding space \\

\hline
$P_{h,i} \subset E_{h,i},\ P_{h,i} \in \mathcal{P}_h$ & Purpose $P_{h,i}$, purpose set $\mathcal{P}_h$\\

\hline
$\iota_{h,i}$ & Purpose intention flag\\
$P_{h,i}^\iota \in \mathcal{P}_{h}^\iota \subseteq \mathcal{P}_{h}$ & Purpose intention $P_{h,i}^\iota$, purpose intention set $\mathcal{P}_{h}^\iota$\\
$d_{h,i}^\iota \in D_{h,i}^\iota \subseteq D$ & Domain intention $d_{h,i}^\iota$, domain intention set $D_{h,i}^\iota$ for $i$\\
$\alpha_{h,i}$ & Priority of purpose\\

\hline
$o_h \in O_h$ & Observation point $o_h$ of observation space $O_h$\\

\hline
$f_h^{S-O}: S_d \to O_h\ \text{or}\ \Delta(O_h)$ & State-observation function: deterministic or stochastic\\
$f_{h}^{O-S}: O_{h} \to \mathscr{P}(S_d) \ \text{or}\ \Delta(S_d)$ & Observation-state relation: multi-valued or stochastic\\
$f_{h,i,d}^{O-U}: O_{h} \to U_{h,i,d}^O$ & Utility function over observations, inherited from $f_{h,i}^{E-U}$\\

\hline
$g_{h,i,d} \in G_{h,i,d} \subset O_h$ & Goal point $g_{h,i,d}$ of goal $G_{h,i,d}$\\
$G_{h,i,d} \in \mathcal{G}_{h,i} \subseteq \mathcal{G}_{h}$, $\mathcal{G}_{h,d} \subseteq \mathcal{G}_{h}$ & Goal sets: $\mathcal{G}_{h,i}$ of purpose $i$; $\mathcal{G}_{h,d}$ in domain $d$; $\mathcal{G}_{h}$ all\\
$G_{h,i,d}^\iota \in \mathcal{G}_{h,i}^\iota \subseteq \mathcal{G}_{h}^\iota$, $\mathcal{G}_{h,d}^\iota \subseteq \mathcal{G}_{h}^\iota$ & Goal intention $G_{h,i,d}^\iota$; goal intention sets: $\mathcal{G}_{h,i}^\iota$, $\mathcal{G}_{h,d}^\iota$,
$\mathcal{G}_{h}^\iota$\\

\hline
$s_{h,i,d} \in S_{h,i,d} \subset S_d$ & Human state goal point $s_{h,i,d}$ and state goal $S_{h,i,d}$\\

\hline
$f^{E_h-E_c}_{h,c,i}: E_h \to E_c$ & Human-robot encoding-space alignment function\\

\hline
\end{tabular}
\end{table*}

\begin{table*}[htb!]
\centering
\caption{Symbols encoding robot related elements, grouped by typology.}
\label{Tab:Symbols3}
\begin{tabular}{ll}

\hline
\textbf{Symbol} & \textbf{Description} \\

\hline
\textit{Robot controller} & \\

\hline
$e_{c,i} \in E_{c,i}$ & Encoding point $e_{c,i}$ of encoding space $E_{c,i}$\\

\hline
$f_{c,i,d}^{O-E}: O_c \to E_{c,i}$ & Observation-encoding function\\
$f_{c,i,d}^{E-O}: E_{c,i} \to \mathscr{P}(O_c)$ & Encoding-observation relation (multi-value function)\\
$f_{c,i}^{E-U}: E_{c,i} \to U_{c,i}^{E}$ & Utility function on encoding, inherited from $f_{h,i}^{E-U}$ \\

\hline
$P_{c,i} \subset E_{c,i},\ P_{c,i} \in \mathcal{P}_c$ & Purpose $P_{c,i}$, purpose set $\mathcal{P}_c$\\
$N_{c,i}\in \mathcal{P}_c$, $M_{c,i} \in \mathcal{P}_c$ & Hardwired need $N_{c,i}$, learnt mission $M_{c,i}$\\

\hline
$\iota_{c,i}$ & Purpose intention flag\\
$P_{c,i}^\iota \in \mathcal{P}_{c}^\iota \subseteq \mathcal{P}_{c}$ & Purpose intention $P_{c,i}^\iota$, purpose intention set $\mathcal{P}_{c}^\iota$\\
$d_{c,i}^\iota \in D_{c,i}^\iota \subseteq D$ & Domain intention $d_{c,i}^\iota$, domain intention set $D_{c,i}^\iota$\\
$\alpha_{c,i}$ & Priority of purpose\\

\hline
$o_c \in O_c$ & Observation point $o_c$ of observation space $O_c$\\

\hline
$f_c^{S-O}: S_d \to O_c\ \text{or}\ \Delta(O_c)$ & State-observation function: deterministic or stochastic\\
$f_{c}^{O-S}: O_{c} \to \mathscr{P}(S_d) \ \text{or}\ \Delta(S_d)$ & Observation-state relation: multi-valued or stochastic\\
$f_{c,i,d}^{O-U} : O_{c} \to U_{h,i,d}^{O}$ & Utility function on observations, inherited from $f_{c,i}^{E-U}$ \\

\hline
$g_{c,i,d} \in G_{c,i,d} \subset O_c$ & Goal point $g_{c,i,d}$ of goal $G_{c,i,d}$\\
$G_{c,i,d} \in \mathcal{G}_{c,i} \subseteq \mathcal{G}_{c}$, $\mathcal{G}_{c,d} \subseteq \mathcal{G}_{c}$ & Goal sets: $\mathcal{G}_{c,i}$ of purpose $i$; $\mathcal{G}_{c,d}$ in domain $d$; $\mathcal{G}_{c}$ all\\
$G_{c,i,d}^\iota \in \mathcal{G}_{c,i}^\iota \subseteq \mathcal{G}_{c}^\iota$, $\mathcal{G}_{c,d}^\iota \subseteq \mathcal{G}_{c}^\iota$ & Goal intention $G_{c,i}^\iota$; goal intention sets: $\mathcal{G}_{c,i}^\iota$, $\mathcal{G}_{c,d}^\iota$, $\mathcal{G}_{c}^\iota$\\

\hline
$s_{c,i,d} \in S_{c,i,d} \subset S_d$ & Robot state goal point $s_{c,i,d}$ of state goal $S_{c,i,d}$\\

\hline
$\mu_{c} \in \mathcal{M}_{c}$, $\mathcal{M}_c = \prod_i E_{c,i}$ & Motivational space point $\mu_c$ and space $\mathcal{M}_c$\\
$f^{\mathcal{M}-U}_c: \mathcal{M} \to U_{c}^\mathcal{M}$ & Utility function over the motivational space \\
\hline
$f^{E_c-E_h}_{h,c,i}: E_c \to E_h$  & Robot-human encoding-space alignment function\\

\hline
\end{tabular}
\end{table*}

\paragraph{\textbf{Core elements}}
Different \textit{humans} (designers/users) are indexed by $h \in H$;  
different \textit{robots} (or \textit{cobots}, collaborative robots) by $c \in C$;  
different \textit{domains} by $d \in D$.

\paragraph{\textbf{Domains}}

Each domain $d \in D$ is characterised by a set of states $s_d \in S_d$, where $S_d$ is the \textit{domain state space}.  
The domain dynamics is governed by a \textit{domain transition function}, which in the deterministic case is defined as:
\[
  f^{SA-S}_{d,c}: S_d \times A_c \to S_d,
\]
and in the stochastic case as:
\[
  f^{SA-S}_{d,c}: S_d \times A_c \to \Delta(S_d),
\]
specifying the deterministic or probabilistic transition from state $s_{d,t}$ to state $s_{d,t+1}$ under the execution of action $a_{c,t} \in A_c$.

\paragraph{\textbf{Human encoding spaces and purposes}} 
Each human $h$ possesses multiple \textit{encoding spaces} $E_{h,i}$, indexed by $i$, each corresponding to a different purpose and comprising points $e_{h,i} \in E_{h,i}$.

A \textit{purpose} $P_{h,i} \subset E_{h,i}$ is defined as:
\[
P_{h,i} = \{ e_{h,i} \in E_{h,i} \mid f_{h,i}^{E-U}(e_{h,i}) \neq 0 \},
\]
where $f_{h,i}^{E-U}: E_{h,i} \to U^E_{h,i} \subseteq \mathbb{R}$ is the \textit{purpose utility function}.
Humans may have multiple purposes $P_{h,i} \in \mathcal{P}_h$.
\textit{Prescriptive} (desired) purposes, essential for fulfilling human intentions, can be defined to encode desired outcomes associated with \textit{positive utilities}.  
Conversely, \textit{proscriptive} (avoidance) purposes, critical for ensuring safety and ethical alignment, can be defined to encode undesired outcomes associated with \textit{negative utilities}.

\paragraph{\textbf{Human observation space and goals}}  
Each human $h$ has \textit{observations} $o_h \in O_h$, forming an \textit{observation space} $O_h$, and a \textit{human observation–encoding function}:
\[
f_{h,i,d}^{O-E}: O_h \to E_{h,i}.
\]
For both humans and robots, for simplicity we assume that the observation–encoding function, and more generally internal processes, are deterministic.

It is useful to define not only the bottom-up functions linking elements of the framework, but also their top-down inverse relations. Since the bottom-up functions are generally non-injective, these inverses are multivalued functions. Such top-down mappings are essential for modelling scenarios in which the robot selects a representation at a higher level, for instance, a purpose point, and uses it to guide the selection of corresponding representations at lower levels, say the associated goal formed by multiple points.
The inverse multivalued \textit{encoding-observation function} --which is domain dependent-- of the function $f_{h,i,d}^{O-E}$, can be formalised as:
\[
f_{h,i,d}^{E-O}:E_{h,i} \to \mathscr{P}(O_h),\quad e_{h,i} \mapsto f_{h,i,d}^{E-O}(e_{h,i}) \subset O_{h} 
\]

where $\mathscr{P}(O_h)$ denotes the powerset of $O_h$ and the function maps a point in the encoding space, $e_{h,i}$, to a subset of elements in $O_h$, that is, $f^{E-O}_{h,i,d}(e_{h,i}) \subset O_h$.

Each purpose point $p_{h,i} \in P_{h,i}$ corresponds to a \textit{human goal} $G_{h,i,d}$ in each domain $d$, defined as:
\[
G_{h,i,d} = \{ o_h \in O_h \mid f_{h,i,d}^{O-E}(o_h) = p_{h,i} \},
\]
or equivalently:
\[
G_{h,i,d} = f^{E-O}_{h,i,d}(p_{h,i}).
\]
Different purpose points $p_{h,i,j} \in P_{h,i}$ generate distinct purpose-related goals $G_{h,i,d,j}$.  
Different purposes $i$ generate different goals in each domain $d$, $\mathcal{G}_{h,d}= \{G_{h,i,d} \mid i=1,2,\dots\}$. 
A purpose point $p_{h,i}$ generates different goals in different domains $D'\subseteq D$ that form the set $\mathcal{G}_{h,i} = \{G_{h,i,d} \mid d\in D'\}$.
All goals of the human form the set $\mathcal{G}_{h}=\{G_{h,i,d} \mid i=1,2,\dots,\; d\in D'\}$. 

Each observation inherits the purpose utility:
\[
  f_{h,i,d}^{O-U}(o_{h}) = f_{h,i}^{E-U}(f_{h,i,d}^{O-E}(o_{h})).
\]
As a result, different goals $G_{h,i,d,j}$ corresponding to different purpose points $p_{h,i,j}$ may have the same or different utilities.

\paragraph{\textbf{States and human state goals}}
The observation space is generated through a \textit{state-observation function} $f_h^{S-O}$, which models the operation of the human sensory system.  
In the deterministic case, it is defined as:
\[
  f_h^{S-O}: S_d \to O_h.
\]
Under a probabilistic assumption, it becomes:
\[
  f_h^{S-O}: S_d \to \Delta(O_h),
\]
where $\Delta(O_h)$ denotes a probability simplex onto the observation space $O_h$.
This function depends solely on the (specific) human sensory apparatus and is therefore independent of the specific domain $d$ and the intended purpose $i$.

Since function $f_h^{S-O}$ commonly abstracts over $S_h$ (partial observability), it is not injective, thus it generates an inverse multivalued \textit{human observation-state function}:
\[
f_{h}^{O-S}: O_{h} \to \mathscr{P}(S_d), \quad o_h \mapsto f_{h}^{O-S}(o_h) \subset S_{d}.
\]
mapping each observation to the set of states that might have generated it.

A point $o_h$ might univocally identify a set of states in one specific domain $d$, $f(o_h) \subset S_d$.
Alternatively, it might correspond to the states of a subset $D' \subseteq D$ of multiple domains, $f(o_h) \subseteq \bigcup_{D'} S_d$.
In the latter case, the human might confound different domains.
Under a stochastic assumption:
\[
f_{h}^{O-S}: O_{h} \to \Delta(S_d).
\]

Based on such functions, each human goal $G_{h,i,d}$ corresponds to a \textit{human state goal}:
\[
S_{h,i,d} = \{ s_d \in S_d \mid f_h^{S-O}(s_d) \in G_{h,i,d} \}=f_{h}^{O-S}(G_{h,i,d}).
\]
Note that the state goal $S_{h,i,d}$ is generated as the union of multiple sets of states, where each set $f_{h}^{O-S}(g_{h,i,d})$ is associated with a single goal point $g_{h,i,d}$:  
$S_{h,i,d} = \bigcup_{g_{h,i,d} \in G_{h,i,d}} f_{h}^{O-S}(g_{h,i,d})$.

\paragraph{\textbf{Human intentions}}
A human typically maintains multiple purposes $\mathcal{P}_h = \{P_{h,1},\dots,P_{h,i},\dots\}$. 
However, at any given time, only a subset of them are \textit{actively pursued} as intended purposes.
In particular, we define \textit{human purpose intentions} the subset $\mathcal{P}_{h}^\iota \subseteq \mathcal{P}_h$ comprising purposes $P_{h,i}^{\iota}$ that are currently \textit{actively pursued}, that is, those for which an intention flag $\iota_{h,i} = 1$: 
$\mathcal{P}_{h}^\iota =\{P_{h,i} \in \mathcal{P}_{h}\mid \iota_{h,i} = 1 \}$. 
The flag takes a value $0$ for purposes not actively pursued.
Note that in the notation such as $P_{h,i}^\iota$ the symbol $\iota$  is reused as superscript for mnemonic simplicity.
Within a purpose intention $P_{h,i}^{\iota}$, the human might select specific purpose point intentions $p_{h,i}^{\iota} \in P_{h,i}^{\iota}$.

The purpose intention flag value is inherited by the corresponding goals that thus become \textit{human goal intentions}  $G_{h,i,d}^{\iota} \in \mathcal{G}_{h}^\iota \subseteq \mathcal{G}_{h}$.
Goal intentions related to a purpose are denoted as $\mathcal{G}_{h,i}^\iota$ and those related to a domain as $\mathcal{G}_{h,d}^\iota$.

The human also specifies the intention that the robot accomplishes the purpose within one or more specific domains.  
These are referred to as \textit{human domain intentions} and are denoted by $d_{h,i}^{\iota} \in D_{h,i}^{\iota} \subseteq D$, where $D_{h,i}^{\iota}$ is the set of intended domains for purpose fulfilment.

\paragraph{\textbf{Robot encoding spaces and purposes}}
Each robot $c$ has multiple \textit{encoding spaces} $E_{c,i}$, with $e_{c,i} \in E_{c,i}$.

A \textit{robot purpose} $P_{c,i} \subset E_{c,i}$ can be either a hardwired \textit{need} $N_{c,i}$, or
a learnt \textit{mission} $M_{c,i}$.

\textit{Robot needs} are defined based on a utility function $f_{c,i}^{E-U}$ hand-crafted by the robot designer:
\[
N_{c,i} = \{ e_{c,i} \in E_{c,i} \mid f_{c,i}^{E-U}(e_{c,i}) \neq 0 \},
\]
where $f_{c,i}^{E-U}: E_{c,i} \to U^E_{c,i} \subseteq \mathbb{R}$.

\textit{Robot missions} are defined through an \textit{alignment function}:
\[
f^{E_c-E_h}_{h,c,i}: E_{c,i} \to E_{h,i}
\]
such that:
\[
M_{c,i} = \{ m_{c,i} \in E_{c,i} \mid f^{E_c-E_h}_{h,c,i}(m_{c,i}) \in P_{h,i} \}.
\]
The alignment function also defines an \textit{inverse alignment function}, possibly multivalued:
\[
f^{E_{h}-E_{c}}_{h,c,i}: E_{h,i} \to \mathscr{P}(E_{c,i}),\quad e_{h,i} \mapsto f^{E_{h}-E_{c}}_{h,c,i}(e_{h,i})\subset E_{c,i}
\]
mapping each human encoding point to the set of robot encoding points corresponding to it.

Utilities propagate from human ones as:
\[
f_{c,i}^{E-U}(m_{c,i}) = f_{h,i}^{E-U}(f^{E_c-E_h}_{h,c,i}(e_{c,i})).
\]

\paragraph{\textbf{Robot observation space and goals}}
Each robot $c$ has observations $o_c \in O_c$, forming the \textit{observation space} $O_c$, and a \textit{robot observation-encoding function}:
\[
f_{c,i,d}^{O-E}: O_c \to E_{c,i}.
\]
This function defines an inverse multivalued \textit{encoding-observation function}:
\[
f^{E-O}_{c,i,d}: E_{c,i} \to \mathscr{P}(O_c),\quad e_{c,i} \mapsto f^{E-O}_{c,i,d}(e_{c,i}) \subset O_c
\]
mapping each encoding point to the set of observations corresponding to it.

Each robot purpose point $p_{c,i}$ induces a different \textit{robot goal} in each domain $d$:
\[
G_{c,i,d} = \{ o_c \in O_c \mid f_{c,i,d}^{O-E}(o_c) \in P_{c,i} \},
\]
or equivalently:
\[
G_{c,i,d} = f^{O-E}_{c,i,d}(P_{c,i}).
\]
Different purpose points $p_{c,i,j} \in P_{c,i}$ generate distinct purpose-related goals $G_{c,i,d,j}$.  
Furthermore, other encoding points $e_{c,i,j} \in E_{c,i}$ may induce goals $G_{c,i,d,j}$ that are instrumental in achieving purpose-related goals.

Goal points inherit utility for purpose points:
\[
f_{c,i,d}^{O-U}(g_{c,i}) = f_{c,i}^{E-U}(f_{c,i,d}^{O-E}(g_{c,i})).
\]

Similarly to humans, robot goals can be grouped by domain, $\mathcal{G}_{c,i}$, by purpose, $\mathcal{G}_{c,d}$, or as a whole set, $\mathcal{G}_{c}$. 

\paragraph{\textbf{States and robot state goals}}  
The observation space is generated through a \textit{robot state–observation function}, which in the deterministic case is defined as:
\[
f_c^{S-O}: S_d \to O_c,
\]
and under a stochastic assumption becomes:
\[
f_c^{S-O}: S_d \to \Delta(O_c).
\]

This function depends on the robot sensory apparatus and here for simplicity it is assumed to be independent of the specific domain $d$ and the intended purpose $i$.  
However, the assumption should be revised if different domains involve different robot sensory systems, as the observation function may then vary accordingly.

This function induces an inverse multivalued \textit{robot observation–state function}:
\[
f_c^{O-S}: O_c \to \mathscr{P}(S_d), \quad o_c \mapsto f_c^{O-S}(o_c) \subset S_d,
\]
which maps each observation to the set of possible states that could have generated it.

For the human, a point $o_c$ may unambiguously identify a set of states within a specific domain $d$, that is, $f(o_c) \subseteq S_d$, or it may correspond to states in a subset $D' \subseteq D$ of multiple domains, $f(o_c) \subseteq \bigcup_{D'} S_d$.  
In the latter case, the robot may be unable to disambiguate the intended domain.  
To resolve such ambiguity, the robot should be provided with explicit information on the domain(s) in which to fulfil the intended purposes.
For simplicity, we will assume this solution in the rest of this work.
Under a stochastic assumption:
\[
f_{c}^{O-S}: O_{c} \to \Delta(S_d).
\]

Importantly, based on such functions, each robot goal $G_{c,i,d}$ corresponds to a \textit{robot state goal}:
\[
  S_{c,i,d} = \{ s_d \in S_d \mid f_c^{S-O}(s_d) \in G_{c,i,d} \}=f_{c}^{O-S}(G_{c,i,d}).
\]
or, under a stochastic assumption:
\[
  f_{c}^{O-S}: G_{c,i,d} \to \Delta(S_d).
\]
In general, when a robot pursues a goal $g_{c,i,d}$, the resulting effects $s_d \in S_d$ are stochastic. However, the deterministic case is useful to represent conditions in which the probability that the robot actions produce unwanted effects --that is, outcomes falling outside a designated set $S_d^\omega := f_{c}^{O-S}(g_{c,i,d})$-- is sufficiently low to be neglected under the operational, safety and ethical constraints considered.

\paragraph{\textbf{Robot intentions}}
As for humans, \textit{robot purpose intentions} correspond to actively pursued purposes, $\mathcal{P}_{c}^\iota =\{P_{c,i} \in \mathcal{P}_{c}\mid \iota_{c,i} = 1\}$, thus $\mathcal{P}_{c}^\iota 
\subseteq \mathcal{P}_{c}$.
The purpose intention flag $\iota_{c,i}$ is inherited by the corresponding goals that thus become \textit{robot goal intentions},  $\mathcal{G}_{c}^\iota \subseteq \mathcal{G}_{c}$.
Goal intentions related to a purpose are denoted as $\mathcal{G}_{c,i}^\iota$ and those related to a domain as $\mathcal{G}_{c,d}^\iota$.
The robot should also have \textit{robot domain intentions} $D_{c,i}^{\iota}$ related to where to perform each purpose: $d_{c,i}^{\iota} \in D_{c,i}^{\iota} \subseteq D$.

\paragraph{\textbf{Ground-truth purposes, goals and state goals}}
Thus far, we have referred to human purposes, goals, state-goals, robot goals, purposes, and the functions mapping between them as \textit{actual elements} characterising the human and the robot.  
These are subject to limitations such as noise, partial observability, learning errors, and misinterpretations.  
However, to analyse the conditions that guarantee alignment, it is crucial to distinguish \textit{actual elements} from \textit{ground truth elements}, which we denote with a superscript $^*$.  
$P_{h,i}^*$ denotes a human purpose that truly reflects the person’s ultimate will and/or aligns with higher-level normative standards (e.g., ethical or societal).  
$G_{h,i,d}^*$ and $S_{h,i,d}^*$ are the goal and state goal corresponding to $p_{h,i}^*$, representing respectively a psychologically grounded mapping, and possible actual states of the world, assumed valid under rational or normative criteria. 

For example, $P_{h,i}^*$ might be verbally expressed as ‘some fruit in a container’, with $p_{h,i}^*$ specified as ‘three fruit items in a kitchen container’ --both reflecting the user’s true intention.  
$G_{h,i,d}^*$ could be a mental image of three apples in a bowl and $S_{h,i,d}^*$ the set of states in domain $d$ that the user would agree to fulfil the purpose.
On the robot side, $G_{c,i,d}^*$ is a suitable perceptual representation corresponding to $S_{h,i,d}^*$, and $p_{c,i}^*$ an abstract representation of $G_{c,i,d}^*$, both defined according to idealised criteria --for example, deemed ‘aligned’ by a putative human expert jury.

\paragraph{\textbf{Purpose priorities and motivational space}}

Robots having multiple intention purposes $\mathcal{P}_c^\iota$ should have mechanisms to arbitrate between them. 
Possible mechanisms usable for this aim are considered in Section \ref{Sec:TheChallengesOfThePurposeFramework}.

We shall see that a means to do so is to attribute \textit{priorities} $\alpha_{c,i}$ to the different purposes and to suitably integrate them for the arbitration.
These priorities are inherited by the related goals.

A second means is the definition of a \textit{motivational space}, based on the Cartesian product of the dimensions of the purposes, which can be used to aggregate the robot multiple purposes:
\[
\mathcal{M}_c = E_{c,1} \times E_{c,2} \times \cdots \times E_{c,n} = \prod_i E_{c,i},
\]
and associate a utility to it:
\[
f^{\mathcal{M}-U}_c: \mathcal{M}_c \to U^\mathcal{M}_c \subseteq \mathbb{R}.
\]
The component of the utility linked to the dimensions of a purpose could possibly be weighted by the purpose priority $\alpha_{h,i}$. 

\paragraph{\textbf{Alignment and triangular alignment}}  
\label{Sec:TriangularAlignment}

\textit{Alignment} occurs when there is a \textit{suitable correspondence} between a human purpose and the corresponding robot purpose.  
On closer analysis, this correspondence should be grounded in what we term \textit{triangular alignment}, which occurs when the human and robot state goals --derived from their respective purposes via encoding-observation and observation-state functions-- coincide.  
Thus, `triangular' refers in particular to the human purpose, the robot purpose, and their shared grounding in observable environment state goals.

The formal conditions under which triangular alignment holds in specific scenarios are investigated in detail in Section~\ref{Sec:AlignmentConditions}.

\paragraph{\textbf{Goal-conditioned reinforcement learning (GC-RL )}}

We assume the robot pursuit of goals, and hence its architecture functioning and behaviour are organised around goals and trials possibly framed as hierarchical goal-conditioned RL problems.

We frame this problem as a \textit{Goal-Conditioned Markov Decision Process} (GC-MDP)~\cite{Santucci2016,LiuZhuZhang2022GoalConditionedReinforcementLearningProblemsAndSolutionsIJCAIVersion,RomeroBaldassarreDuroSantucci2023AutonomousOpenEndedLearningofTaskswithNonStationaryInterdependencies}.
A GC-MDP can be defined by the tuple $(S_d, A_c, O_c, G_{c,i,d},   f_{d,c}^{SA-S}, R_c, \gamma)$.
Here the additional symbol $\gamma$ indicates a discount factor and $R(s_{t}, a_{t}, s_{t+1}, G_{c,i,d})$ a goal-conditioned reward function.

For a given purpose-satisfying goal $G_{c,i,d}$, we denote by $G_{c,i,d,j}$ the instrumental subgoals selected to pursue it and forming a sequence indexed with $j=1,2,\cdots,n$.
Such subgoals might be several, one, or none.

We denote by $t_j$ the first time step of the pursuit of $G_{c,i,d,j}$, and by $\tau_{j}$ the number of time steps, and primitive actions $a_{c,t}$, taken to accomplish it.
We denote by
$Z_{c,0\_t}^{OA} = \langle o_{c,0},\,a_{c,0},\,\dots,\,o_{c,t},\,a_{c,t}\rangle$ the robot $c$ observation action history from $t=0$ to $t$, and by 
$Z_{c,t_j\_\tau_j}^{OA} = \langle  o_{c,t_j},\,a_{c,t_j},\,\dots,o_{c,t_j+\tau_j} a_{c,t_j+\tau_j}\rangle$ the robot observation action history from $t_j$ to $t_j+\tau_j$
to accomplish $G_{c,i,d,j}$.
Analogously, we denote by
$Z_{c,0\_t}^{SA} = \langle s_{c,0},\,a_{c,0},\,\dots,s_{c,t},\,a_{c,t}\rangle$ the domain $d$ state action history until $t$, and with
$Z_{c,t_j\_\tau_j}^{SA} = \langle s_{c,t_j},\,a_{c,t_j},\,\dots,s_{c,t_j+\tau_j},a_{c,t_j+\tau_j}\rangle$ the domain state action history for $G_{c,i,d,j}$.
We denote the histories $Z_{c,0\_t}^{A}$, $Z_{c,t_j\_\tau_j}^{A}$, $Z_{d,0\_t}^{S}$, $Z_{c,t_j\_\tau_j}^{S}$ analogously.

We introduce two action policies.
The first is a high-level policy $\Pi_c$ (\textit{goal selector}) that selects a subgoal $G_{c,i,d,j} \sim \Pi_c(\cdot \mid Z_{c,0\_t}^{OA},G_{c,i,d})$ based on the observation-action history $Z_{c,0\_t}^{OA}$ and goal $G_{c,i,d}$.

The second is a low-level policy $\pi_c$ (\textit{goal-pursuer}) that selects an action $a_{c,t} \sim \pi_c(\cdot| Z_{c,0\_t}^{OA}, G_{c,i,d,j})$ based on the history $Z_{c,0\_t}^{OA}$ and the subgoal $G_{c,i,d,j}$.

High-level decisions occur on a slower time scale, providing temporal abstraction.
Low-level decisions occur at every time step, selecting actions to pursue the currently active goal.

We use the \textit{enabledness predicate} $\mathcal{E}(\cdot)$ to indicate that, conditioned on an event such as the accomplishment of another subgoal $G_{c,i,d,j}$, a given subgoal of interest $G_{c,i,d,j+1}$ becomes possible to pursue: $G_{c,i,d,j} \rightarrow \mathcal{E}(G_{c,i,d,j+1})$.  

As we shall see, on this basis we can consider situations where, given the initial state $s_{d,t}$, the robot is able to accomplish the goal $G_{c,i,d}$ by (a) selecting a suitable sequence of subgoals $\langle G_{c,i,d,1},\, G_{c,i,d,2},\, \cdots\,, G_{c,i,d,n},\, G_{c,i,d} \rangle$ through the goal-selection policy $G_{c,i,d,j} \sim \Pi_c(\cdot \mid Z_{c,0\_t}^{OA}, G_{c,i,d})$, and (b) selecting a suitable sequence of actions $Z_{c,t_j\_\tau_j}^{A}$ through the action policy $a_{c,t} \sim \pi_c(\cdot \mid Z_{c,0\_t}^{OA}, G_{c,i,d,j})$ to accomplish each $G_{c,i,d,j}$.

The reward function $R(s_{t}, a_{t}, s_{t+1}, G_{c,i,d,j})$ can be used to mark the successful achievement of subgoal $G_{c,i,d,j}$ within a certain \textit{trial timeout} $\tau \geq \tau_{j}$, when the reward exceeds a given threshold:\\
$R(s_{t}, a_{t}, s_{t+1}, G_{c,i,d,j}) > Th^G$.

\section{Taxonomy of purposes}
\label{Sec:PurposeClasses}

\subsection{Primitive and learnt robot purposes}

We have seen that two main classes of robot purposes can be distinguished, based on their origin: \textit{primitive needs} and \textit{learnt missions}.
These are now analysed in more detail.

\paragraph{Primitive robot purposes: needs}
Primitive purposes, or \textit{needs}, are hardwired by the designer prior to the robot's deployment.

A first category consists of \textit{implicit needs}, embedded in hardwired algorithms of the robot architecture (e.g., obstacle avoidance reflexes).

A second category includes \textit{explicit needs}, represented within robot encoding spaces $E_{c,i}$ and associated with a utility function
$f_{c,i}^{E-U}: E_{c,i} \to U_{c,i}^{E}$.
Explicit needs define utility-bearing subsets $N_{c,i} \subset E_{c,i}$:
$
N_{c,i} = \{ e_{c,i} \in E_{c,i} \mid f_{c,i}^{E-U}(e_{c,i}) \neq 0 \}
$.
An example is the need to maintain a high battery level.

Importantly, certain needs can instantiate \textit{meta-purposes}, such as driving the robot to interact with users to acquire missions and associated utility/prioritisation structures.

Finally, explicit or implicit needs can encode \textit{ethical and safety constraints}, often assigned a high priority over operational missions to ensure compliance with human-centred values.

\paragraph{learnt robot purposes: missions}
Missions are purposes that the robot autonomously acquires during its operational life.

A first class of missions involves the \textit{encoding of human purposes} learnt via explicit interactions (e.g., verbal instructions, demonstrations).

A second class concerns \textit{instrumental self-generated missions}, created to support the fulfilment of preexisting needs or missions.
These involve the autonomous generation of:
(a) a new encoding spaces $E_{c,i}$; (b) a new missions $M_{c,i} \subset E_{c,i}$; (c) the mission utility functions $f_{c,i}^{E-U}$ and possibly priority $\alpha_{c,i}$.

The self-generation of encoding spaces is non-trivial, and should rely on representation learning techniques \cite{BengioCourvilleVincent2013RepresentationLearningaReviewandNewPerspectives}.
A promising approach is to employ highly expressive general-purpose representation spaces, such as language, processed via large language models (LLMs).

\subsection{Taxonomies of purposes related to motivation classes}

Purposes can also be classified according to the traditional taxonomy of motivations in OEL in the literature \cite{Baldassarre2022IntrinsicMotivationsforOpenEndedLearning}, distinguishing \textit{extrinsic} and \textit{intrinsic} motivations.

\paragraph{Extrinsic purposes related to human purposes}
\textit{Extrinsic purposes} aim to produce desired effects in the external physical or social environment.
Among these, \textit{operational extrinsic purposes} involve missions or needs whose goals directly satisfy human purposes through \textit{environmental changes} (e.g., sorting fruits, tidying spaces).
\textit{Social extrinsic purposes} involve \textit{modifying social or psychological states} (e.g., serving food to people, entertaining children).
These purposes are typically associated with encoding spaces $E_{c,i}$ defined over physical, social, or psychological environmental features.

\paragraph{Extrinsic purposes related to homeostatic needs}
A second class of extrinsic purposes addresses \textit{homeostatic needs}, essential for self-preservation and operational continuity.
Examples include maintaining battery charge, protecting mechanical integrity, and ensuring functional sensorimotor capacities.
Such needs are encoded in multidimensional spaces $E_{c,i}$ with a purpose utility function $f_{c,i}^{E-U}(e_{c,i})$ driving maintenance behaviours, for example with $e_{c,i}$ being the battery level and wheel health.

\paragraph{Intrinsic purposes}  
\textit{Intrinsic purposes} are epistemic drives that motivate the robot to acquire new knowledge, skills, or improved models of the world, independent of immediate external objectives.  
For example, a class of competence-based intrinsic purposes might be defined in an encoding space $E_{c,i}$ that represents internal metrics such as skill competence improvement.
The associated utility function $f_{c,i}^{E-U}$ could reward expected improvements in competence across goals:
\[
  f_{c,i}^{E-U}(\mathbb{P}_{G{c,i}}) = \mathbb{E}_{g_{c,i} \sim \mathbb{P}_{G_{c,i}}} \left[ \delta \mathcal{C}(g_{c,i}) \right], \quad \mathbb{P}_{G_{c,i}} \in \Delta(G_{c,i})
\]
where $\mathbb{P}_{G_{c,i}}$ is a probability distribution over the current set of robot goals $G_{c,i}$, and $\delta \mathcal{C}(g_{c,i})$ denotes the expected competence improvement for goal $g_{c,i}$ (e.g., $\delta \mathcal{C}(g_{c,i})) = \mathcal{C}_t(g_{c,i}) - \mathcal{C}_{t-\tau}(g_{c,i})$).

A class of knowledge-based intrinsic purposes may use utility functions that reward reductions in uncertainty or improvements in model accuracy.  
In the latter case, the utility function can be formalised in Bayesian terms as~\cite{BartoMirolliBaldassarre2013Noveltyorsurprise,Friston2015,Taniguchi2023}:  
\[
  f_{c,i}^{E-U}(\mathrm{M}, \mathrm{D}) = D_{\mathrm{KL}}\left( \mathbb{P}(\mathrm{M} \mid \mathrm{D}) \parallel \mathbb{P}(\mathrm{M}) \right),
\]
where $D_{\mathrm{KL}}\left( \cdot \parallel \cdot \right)$ is the Kullback-Leibler divergence, $\mathbb{P}(\mathrm{M})$ is the prior distribution over models $\mathrm{M} \in \mathscr{M}$, and $\mathbb{P}(\mathrm{M} \mid \mathrm{D})$ is the posterior after observing new data $\mathrm{D} \in \mathscr{D}$.

\section{Four important sub-problems of the autonomy-alignment problem}
\label{Sec:TheChallengesOfThePurposeFramework}

The purpose framework highlights four major sub-problems into which the autonomy-alignment problem can be decomposed.
These are discussed below, along with possible strategies for addressing them.

\subsection{Arbitration of purposes problem}
\label{Sec:ArbitrationOfPurposesProblem}

\paragraph{Problem}
How can a robot arbitrate between competing purposes, potentially using \textit{priorities} $\alpha_{c,i}$ or the \textit{motivation space} $\mathcal{M}_{c}$?
In particular, how should the robot weigh the relative importance of multiple purposes when they are either \textit{incompatible} (cannot be pursued simultaneously) or \textit{compatible}?

\paragraph{Strategies toward solutions}

\paragraph{Hierarchical arbitration}

When \textit{purposes are incompatible}, a classical approach is to impose a strict hierarchical structure by assigning fixed priorities to each purpose.
This strategy mirrors arbitration schemes in behaviour-based robotics, such as the \textit{subsumption architecture}~\cite{Brooks1986ARobustLayeredControlSystemforaMobileRobot}, where higher-priority behaviours inhibit or override lower-priority ones.
For example, safety-related purposes might always take precedence over epistemic or operational purposes.
Formally, arbitration selects the purpose with the highest priority among the currently intended (active) ones:
\[
   P_{c,i}^{\text{selected}} = \operatorname*{argmax}_{P_{c,i}} f_{c}^{P^\iota-\mathcal{A}}(P_{c,i}^\iota).
\]
where the function $f_{c}^{P^\iota-\mathcal{A}}: P_{c}^\iota \to \mathcal{A}_{c}$ maps purposes to their priority ($\alpha_{c,i}\in \mathcal{A}_c$).

A more flexible alternative selects purposes dynamically, combining priority weights with urgency metrics based on purpose satisfaction. One such arbitration rule is:
\[
  P_{c,i}^{selected} = \operatorname*{argmax}_{P_{c,i}} 
  \left[ \alpha_{c,i} \left(1 - f_{c,i}^{E-U}(f_{c}^{\mathcal{P}^\iota-E_i}(P_{c,i})) \right) \right],
\]
where $f_{c,i}^{E-U}(e_{c,i})$ denotes the utility (assumed in $(0,1)$) of the currently active encoding point $e_{c,i}$, returned by the function $f_{c}^{\mathcal{P}^\iota-E_i}: \mathcal{P}_{c}^\iota \to E_{c,i}$ mapping an intended purpose $P_{c,i}^{\iota}$ to its related currently-active encoding-space point $e_{c,i}$.
Alternatively, a probabilistic arbitration rule could be used, for instance, via a \textit{softmax} distribution over purposes:
\[
  \mathbb{P}(P_{c,i}) =
  \frac{\exp\left( \frac{\alpha_{c,i} \left(1 - f_{c,i}^{E-U}\left(f_{c}^{\mathcal{P}^\iota-E_i}(P_{c,i})\right) \right)}{\tau} \right)}
  {\sum_j \exp\left( \frac{\alpha_{c,j} \left(1 - f_{c,j}^{E-U}\left(f_{c}^{\mathcal{P}^\iota-E_i}(P_{c,i})\right) \right)}{\tau} \right)}.
\]
where $\tau$ is a temperature parameter controlling exploration.

If \textit{purposes are compatible}, arbitration can rely on the motivation space $\mathcal{M}_c$ (see Section \ref{Sec:FormalisationOfThePurposeFramework}).
A utility function $f^{\mathcal{M}-U}_c$ on the entire motivational space can be defined, for example, by integrating the utility of purposes based on their priorities $\alpha_{c,i}$:
\[
  f^{\mathcal{M}-U}_c(\mu_c) = \sum_{i} \alpha_{c,i} \cdot f_{c,i}^{E-U}(e_{c,i}).
\]

In this context, it is important to assign distinct roles to \textit{priorities} and \textit{utility functions}.

Arbitration between purposes should primarily rely on \textit{priorities} $\alpha_{c,i}$ rather than on altering \textit{utility functions} $f_{c,i}^{E-U}$.
In fact, priorities are suitable for regulating the \textit{relative importance between purposes}, whereas utility functions to establish the \textit{relative desirability of points internal to a purpose}.
Thus, acting on utilities to arbitrate between different purposes would introduce distortions on the relative importance of purpose points.

Priorities can be \textit{static}, not changing in time, or \textit{dynamic}, thus changed during the robot operation based on factors such as the feasibility of purposes, their success rate, or context (e.g., different users, different domains).
Such adjustments should rely on changing priorities $\alpha_{c,i}$ rather than altering utility functions $f_{c,i}^{E-U}$, for the same reasons discussed above.

\subsection{The human–robot alignment problem}
\label{Sec:TheHumanRobotAlignmentProblem}

\paragraph{Problem}
How can we ensure that the effects produced by robot purposes align with those expected by humans, given their intended purposes and domains?

\paragraph{Strategies towards solutions}
In this section, we discuss four major implementation solution strategies, while Section~\ref{Sec:AlignmentConditions} provides a formal analysis of alignment in key conditions.

\begin{itemize}
  \item  \textit{Hardwired needs}. 
  Robot designers could hardwire into the robot the encoding spaces $E_{c,i}$ of needs $N_{c,i}$, their utility functions $f_{c,i}^{E-U}$, and their priorities $\alpha_{c,i}$.
  The main challenge is to ensure that the hardwired needs accurately reflect the corresponding human purposes (see Section~\ref{Sec:AlignmentLiterature}), that is, they lead to the expected effects.
  \item 
  \textit{Top-down mission acquisition via shared encoding spaces}.  
  Humans could assign missions $M_{c,i}$ to the robot through shared general-purpose encoding spaces, $E_{h,i} \approx E_{c,i}$, such as language-based representations.  
  A key challenge lies in aligning the human and robot semantic groundings of purposes, due to the inherent ambiguity and subjectivity of language, which may require extensive human feedback.  
  An additional difficulty is the robot's ability to infer or learn suitable goals that satisfy the acquired purposes.
  \item
  \textit{Bottom-up mission acquisition via robot goal instances}. 
  The robot infers missions $M_{c,i}$ by receiving multiple examples of \textit{robot goal points} $g_{c,i,d} \in G_{c,i,d}$ if the robot observation space $O_{c}$ is known, and such points can be generated knowing that they correspond to the state goal points desired by the human.
  For example, $g_{c,i,d}$ could be examples of a dataset of images corresponding to the robot camera.
  Challenges include the construction of the dataset and the inference of the abstract purpose $P_{c,i}$ based on examples $g_{c,i,d}$.
  \item
  \textit{Bottom-up mission acquisition via domain goal instances}. 
  The robot infers missions $M_{c,i}$ by observing multiple examples of human-satisfying \textit{state goals} $s_{h,i,d} \in S_{h,i,d}$, or autonomously discovers goals under human validation.
  Challenges include the feedback cost for the user and the robot generalisation from finite experience samples to broader goal and purpose representations.
\end{itemize}

\subsection{The purpose-goal grounding problem}
\label{Sec:ThePurposeGoalGroundingProblem}

\paragraph{Problem}
Purposes are defined in domain-independent encoding spaces, whereas goals must be instantiated within specific domains.
How can the robot map abstract purposes to domain-specific goals corresponding to concrete domain states?

\paragraph{Strategies towards solutions}
There are multiple ways in which a robot might ground a purpose into specific goals.  
To illustrate this, we briefly describe a strategy currently under development~\cite{CartoniCioccoliniBaldassarre2025FocusingRobotOpenEndedReinforcementLearningThroughUsersPurposes}:
\begin{itemize}
  \item The robot receives a purpose in abstract form, such as a linguistic statement (e.g., `put some fruit in a container').
  \item It identifies and localises objects in a designated target domain via segmentation and object recognition.
  \item It selects purpose-relevant objects, such as `bananas' and `bowl'.
  \item It interacts with the domain to produce and learn goal states that satisfy the abstract purpose (e.g., `banana in bowls').
\end{itemize}
The grounding process relies on multiple machine learning components, including natural language processing, visual perception and scene understanding, motor control, and autonomous learning.

\subsection{The competence acquisition problem}
\label{Sec:TheCompetenceAcquisitionProblem}

\paragraph{Problem}
How can the robot learn to achieve the grounded goals associated with purposes?

\paragraph{Strategies towards solutions}
The robot can learn policies using reward functions $R(o_{c,t}, a_{c,t}, o_{c,t+1}, G_{c,i,d})$, associating rewards with successful transitions towards goals.

Within robotics, utility functions $f_{c,i}^{E-U}$ tend to be used for evaluating \textit{static desirability} over observations or encodings, while reward functions $R$ to assess \textit{state transitions}, thus possibly serving complementary functions \cite{HubingerMerwijkMikulikSkalseGarrabrant2019RisksFromLearnedOptimizationInAdvancedMachineLearningSystems}.
However, goals $G_{c,i,d}$, defined as desirable subsets of observations based on utility functions $f_{c,i,d}^{O-U}$, allow the generation of \textit{pseudo-reward functions} (e.g., see \cite{SinghBartoChentanez2004IntrinsicallyMotivatedReinforcementLearning}):
\[
  R(o_{c,i,d}) = 
   \begin{cases}
     1 & \text{if } o_{c,i,d} \in G_{c,i,d},\\
     0 & \text{otherwise}.
   \end{cases}
\]

The robot may further estimate \textit{expected utility functions} within its internal spaces, similarly to the \textit{evaluation functions} associated to reward functions within the RL framework:

\begin{itemize}

\item \textit{Observation space} $O_c$.
In this case, the expected utility function related to a goal, covering the entire space $O_c$, can be defined on the basis of the utility of robot goals similarly to what is done in RL.
This expected utility might be used to guide policy learning via desirability gradients, in particular when goals are formed by relatively few points.

\item \textit{Encoding space} $E_{c,i}$. 
In this case, one could define the expected utility functions based on metrics that weight the discrepancy of the currently active encoding point from the points that make up the purpose.
However, such metrics should suitably change depending on the different domains, as these would be abstracted differently by the encoding space.
Also these types of expected utilities might be used to guide policy learning via desirability gradients.

\end{itemize}

These expected utility functions could be either hardwired or learnt through RL techniques \cite{Sutton1998}, possibly with initial shaping based on prior knowledge.

\section{Formal definition of alignment, and necessary and sufficient conditions, in key conditions}
\label{Sec:AlignmentConditions}

This section initiates the investigation of necessary and sufficient conditions for \textit{alignment} within the formal framework.
As we have seen in Section~\ref{Sec:FormalisationOfThePurposeFramework}, \textit{alignment} actually involves a triadic relation --\textit{triangular alignment}-- that involves
(a) the human purpose,
(b) the robot purpose, and
(c) the corresponding state goals in the considered domain.
Broadly speaking, the definitions of alignment proposed here involve the relation `a-b' involving the aforementioned three elements, while the necessary and sufficient conditions refer to the relations `a-c' and `b-c'.

We assume that human intentions toward the robot are expressed as explicit \textit{purposes}, corresponding to the highest representational level defined in the framework.
However, in reality psychological research shows that human goals span multiple levels of abstraction, including habitual systems that operate with limited or no conscious awareness (e.g., \cite{Kahneman2011ThinkingFastAndSlow}).
As a result, humans may expect robots to perform behaviours linked to lower-level, implicit, or even internally inconsistent objectives.
Although these complexities are significant, they fall outside the scope of this work.

In the following, we also make a simplifying assumption: when a human selects a purpose, this is not accomplished unless the robot actively attempts to satisfy it.
That is, we assume that the purpose is not fulfilled by the environment or by the action of another agent in the absence of the robot's intervention.
This avoids addressing alignment under additional conditions where the robot needs not act directly but merely intends to pursue the purpose and monitors external events or agents for its satisfaction.
In \ref{Sec:Pearl}, we analyse the conditions under which the robot action constitutes the actual \textit{cause} of the fulfilment of human purposes, based on Pearl's theory of causality \cite{HalpernPearl2005CausesAndExplanationsAStructuralModelApproachPartICauses,Pearl2009CausalityModelsReasoningAndInference}.

Here we focus on the conditions for alignment by analysing these increasingly complex cases:
(a) alignment with an \textit{extrinsic purpose} in a single domain;
(b) alignment with an \textit{extrinsic purpose} with \textit{variable utility} points in a single domain;
(c) alignment with an \textit{intrinsic purpose} in a single domain;
(d) alignment with an \textit{extrinsic purpose} and the capacity to pursue \textit{instrumental goals} beyond the intrinsic purpose within a single domain;
(e) alignment with an \textit{extrinsic purpose}, the pursuit of \textit{instrumental goals}, and the presence of a high-priority \textit{proscriptive extrinsic purpose} (e.g, a safety constraint) in a single domain;
(f) alignment across \textit{multiple domains}.
Further research is needed to address more complex scenarios, such as alignment involving multiple interacting purposes with diverse arbitration mechanisms.
These extensions are left to future work.

\subsection{Alignment with an extrinsic purpose}
\label{Sec:ExtrinsicPurposeProof}

In this section, we consider the case where the human intends the robot to fulfil a single extrinsic purpose within a specified domain. 
In this and all the following conditions, we assume that, after being suitably informed about human intention, the robot formulates and acts according to a \textit{robot intention tuple} 
$\langle P_{c,i}^\iota,\ p_{c,i}^\iota,\ d_{c,i}^\iota\rangle$

comprising:
(a) a purpose $P_{c,i}^\iota$;  
(b) a point $p_{c,i}^\iota$ within that purpose;  
(c) a domain $d_{c,i}^\iota$ in which to pursue it.  

For simplicity of notation, as it is often used as index, we denote the human-intended domain $d_{h,i}^{\iota*}$ as $d$.

Alignment can hence be defined as follows.

\noindent\textbf{Definition: alignment with an extrinsic purpose.}  
\textit{When the human has an extrinsic purpose composed of points with uniform utility, human-robot alignment with respect to purpose $P_{h,i}^{\iota*}$ and domain $d$ holds if:}

\begin{equation}
  \begin{aligned}
  &\left(\mathcal{P}_{h}^{\iota*} = \{P_{h,i}^{\iota*}\} \;\land\; D_{h,i}^{\iota*} = \{d\}\right) 
  \rightarrow \\
  &\exists\, P_{c,i}^{\iota*} \;
  \exists\, p_{c,i}^{\iota*} \;
  \exists\, d_{c,i}^{\iota*} 
  \left(p_{c,i}^{\iota*} \in P_{c,i}^{\iota*} \;\land\; 
  d_{c,i}^{\iota*} = d \;\land\; 
  e_{h,i}^\omega \in P_{h,i}^{\iota*}\right)
 \end{aligned}
 \label{Eq:DefinitionAlignmentExtrinsic}
\end{equation}  

\textit{where  
$P_{h,i}^{\iota*}$ and $d$ denote the human’s intended purpose and target domain;  
$P_{c,i}^{\iota*}$, $p_{c,i}^{\iota*}$, and $d_{c,i}^{\iota*}$ are the robot selected purpose, purpose point, and domain of execution;  
$e_{h,i}^\omega$ denotes the human's encoding of the outcome of the robot's action, caused by pursuing $p_{c,i}^{\iota*} \in P_{c,i}^{\iota*}$.}

The definition states that alignment holds when the following occurs.  
If the human desires that purpose $P_{h,i}^{\iota*}$ be accomplished in domain $d$, the robot selects --and forms, if not already present-- a corresponding purpose $P_{c,i}^{\iota*}$, chooses a purpose point $p_{c,i}^{\iota*} \in P_{c,i}^{\iota*}$, and selects domain $d_{c,i}^{\iota*} = d$.
The key element of the definition is that the robot action causes an environment state such that the human, upon perceiving and classifying the outcome as $e_{h,i}^\omega$ in their abstract encoding space $E_{h,i}$, recognises it as satisfying the intended purpose, that is, $e_{h,i}^\omega \in P_{h,i}^{\iota*}$.

An example that tokenises these symbols is as follows.  
The human purpose $P_{h,i}^{\iota*}$ might correspond to a linguistic description such as `some fruit in a container'.  
The domain $d$ could be a factory floor containing apples and basket containers.  
The robot's purpose $P_{c,i}^{\iota*}$ might be represented by the same linguistic sentence, for instance interpreted through a large language model.  
The purpose point $p_{c,i}^{\iota*}$ could specify a concrete target, such as `three fruits in a container', to be pursued in the domain $d_{c,i}^{\iota*}$ indicated by the human.  
The goal $G_{c,i,d}^{\iota*}$ considered in the theorem below,instantiating $p_{c,i}^{\iota*}$ in the specific $d$, might be `three apples in the green basket'.  
Finally, the abstract outcome $e_{h,i}^\omega$ would correspond to the human interpretation of the environmental effects produced by the robot's actions.

Although abstract, the purpose framework \textit{de facto} presupposes certain features of the robot's internal architecture.  
These assumptions enable a formal characterisation of the necessary and sufficient conditions under which alignment holds, as stated in the following theorem.  
These conditions establish that alignment between human and robot purposes is grounded in the environmental effects caused by the robot's actions once it commits to a purpose and as these effects are perceived and evaluated by the human.  
When this causal grounding is emphasised, we refer to it as \textit{triangular alignment}, rather than simply `alignment'.

\noindent\textbf{Theorem: extrinsic purpose.}
\textit{In the case where the human has an extrinsic purpose composed of points with unfiorm utility, and under the assumptions of the purpose framework, the necessary and sufficient conditions for human-robot alignment are as follows:}

\begin{equation}
 \begin{aligned}
  &(1)\ \exists\, G_{c,i,d}^{\iota*} \subseteq f^{E-O}_{c,i}(p_{c,i}^{\iota*})\; \land \\
  &(2)\ f_{c}^{O-S}(G_{c,i,d}^{\iota*}) = S_{c,i,d}^\omega \subseteq S_{h,i,d}^*  \; \land \\
  &(3)\ f_h^{S-O}(S_{c,i,d}^\omega) = O_{h,i,d}^\omega \subseteq G_{h,i,d}^*\; \land \\
  &(4)\ f^{O-E}_{h,i}(O_{h,i,d}^\omega) = E_{h,i}^{\omega} \subseteq P_{h,i}^{\iota*}
 \end{aligned}
 \label{Eq:TheoremAlignmentExtrinsic}
\end{equation}  

The theorem establishes that human-robot alignment with an extrinsic purpose $P_{h,i}^{\iota*}$ is achieved when the robot selects a goal $G_{c,i,d}^{\iota*}$ corresponding to the purpose point $p_{c,i}^{\iota*}$ (condition 1), and when pursued this goal causally leads to world states $S_{c,i,d}^\omega$ falling within the human-desired environmental states $S_{h,i,d}^*$ (condition 2).
These states must be perceived by the human as observations $O_{h,i,d}^\omega$ that the human recognises as satisfying their goal $G_{h,i,d}^*$ (condition 3), and these observations must also be encoded by the human as purpose-relevant points within their intended purpose, $E_{h,i}^{\omega} \subseteq P_{h,i}^{\iota*}$ (condition 4). 

Together the conditions 1-4 establish that for each purpose $P_{h,i}^{\iota*}$ the human might desire, the robot selects a goal $G_{c,i,d}^{\iota*}$ that ignites a causal chain accomplishing the purpose (sufficiency).
This chain is well represented by the compound function:
\begin{equation}
\forall P_{h,i}^{\iota*}\; \exists G_{c,i,d}^{\iota*}\left(
f_{h,i}^{O-E} \circ 
f_{h}^{S-O} \circ 
f_{c}^{O-S}(G_{c,i,d}^{\iota*})
\subseteq P_{h,i}^{\iota*}\right)
\end{equation}
If any link in this chain is broken --goal selection, environmental effect, perception, or interpretation-- alignment fails (necessity).
The proof of the theorem is presented in \ref{Sec:AppendixExtrinsicPurposeProof}.

\subsection{Alignment with an extrinsic purpose and variable utility}
\label{Sec:ExtrinsicPurposeVariableUtilityProof}

We now consider the case in which the human desired purpose consists of points associated with variable utility, $f_{h,i}^{E-U}(p_{h,i}^{})$.
Two subcases arise under this condition.  
In the first subcase, the robot is required to pursue a purpose point that leads to a human purpose point whose \textit{utility exceeds a specified threshold}.  
For example, the purpose might be `collect some fruit into a container with a maximum volume of 4 litres', with the threshold specified as the minimum number of fruit items to gather.
In the second subcase, the robot must pursue a purpose point that leads to the human purpose point with \textit{maximal utility}.  
For example, the purpose might be `fit the maximum number of fruit items into a container with a maximum volume of 4 litres'.
We begin by focusing on the first subcase.

\noindent\textbf{Definition: alignment with extrinsic purpose and variable utility.}
\textit{When the human has an extrinsic purpose $P_{h,i}^{\iota*}$ formed by points with utility values $f_{h,i}^{E-U}(p_{h,i}^{\iota*})$, human-robot alignment with respect to $P_{h,i}^{\iota*}$ and domain $d$ holds if:}
\begin{equation}
\begin{aligned}
  &\left(\mathcal{P}_{h}^{\iota*} = \{P_{h,i}^{\iota*}\} \;\land\; D_{h,i}^{\iota*} = \{d\}\right) 
  \rightarrow \\
  &\exists\, P_{c,i}^{\iota*} \;
  \exists\, p_{c,i}^{\iota*} \;
  \exists\, d_{c,i}^{\iota*} \\
  &\left(p_{c,i}^{\iota*} \in P_{c,i}^{\iota*} \;\land\; 
  d_{c,i}^{\iota*} = d \;\land\; f_{h,i}^{E-U}(e_{h,i}^\omega) > \theta_{h,i}^{E-U} \right)
\end{aligned}
\label{Eq:AppendixDefinitionAlignmentExtrinsicUtility}
\end{equation}
\textit{where $\theta_{h,i}^{E-U}$ is a task-dependent utility threshold above which the outcome $e_{h,i}^\omega$ is considered to satisfy the human intention.}

The definition is similar to the previous uniform utility case, with the exception of the final requirement $f_{h,i}^{E-U}(e_{h,i}^\omega) > \theta_{h,i}^{E-U}$ demanding that the robot action causes not just the pursuit of any purpose point, but also that this point has a utility above a certain threshold.
In this case, the necessary and sufficient conditions are established by the following theorem.

\noindent\textbf{Theorem: alignment with extrinsic purpose and variable utility.}
\textit{In the case where the human has an extrinsic purpose $P_{h,i}^{\iota*}$ composed of points with associated utilities, and under the assumptions of the purpose framework, the necessary and sufficient conditions for human-robot alignment are:}
\begin{equation}
 \begin{aligned}
  &(1)\ \exists\, G_{c,i,d}^{\iota*} \subseteq f^{E-O}_{c,i}(p_{c,i}^{\iota*})\; \land \\
  &(2)\ f_{c}^{O-S}(G_{c,i,d}^{\iota*}) = S_{c,i,d}^\omega \subseteq S_{h,i,d}^*  \; \land \\
  &(3)\ f_h^{S-O}(S_{c,i,d}^\omega) = O_{h,i,d}^\omega \subseteq G_{h,i,d}^*\; \land \\
  &(4)\ f^{O-E}_{h,i}(O_{h,i,d}^\omega) = e_{h,i}^{\omega} \ \land\ f_{h,i}^{E-U}(e_{h,i}^{\omega}) > \theta_{h,i}^{E-U}
 \end{aligned}
\label{Eq:AppendixTheormeAlignmentExtrinsicUtility}
\end{equation}

The conditions differ from those in the uniform utility case only in the final part of Condition~(4).  
This condition is now strengthened to require that the set of possible human-perceived outcomes $e_{h,i}^{\omega}$ have utility exceeding a given threshold, $f_{h,i}^{E-U}(e_{h,i}^{\omega}) >\theta_{h,i}^{E-U}$.

The second subcase, in which the robot must maximise the utility of the human-perceived purpose point, is captured by replacing the threshold condition in both the definition and the theorem with a maximisation requirement $e_{h,i}^\omega = \arg\max_{e_{h,i} \in P_{h,i}^{\iota*}} f_{h,i}^{E-U}(e_{h,i})$.

Importantly, in both subcases alignment requires that the robot not only learns/is given (at least some of) the purpose points corresponding to those of the human (along with their associated goals), but also the utility function over those points in a way that reflects the utility the human ascribes to their corresponding purpose points.
This poses significant algorithmic challenges.
See \ref{Sec:AppendixExtrinsicPurposeVariableUtilityProof} for the proof of the theorem.

\subsection{Alignment with an intrinsic purpose}
\label{Sec:IntrinsicPurposeProof}

We now consider a case that is closer to typical OEL scenarios.
Intrinsic purposes become critical when the user aims for the robot to autonomously learn a broad range of capabilities within a certain class, while reserving the assignment of specific objectives for a later stage, possibly through specific extrinsic purposes.
The human is satisfied if the robot discovers and learns to accomplish with high competence as many goal points as possible --possibly all-- fulfilling the purpose.
For instance, a user might initially ask the robot to `learn to manipulate apples and apple containers in this room' through OEL, and subsequently assign a specific purpose such as `collect three apples in this green basket' and expect a zero-shot performance.

Alignment can then be defined as follows.

\noindent\textbf{Definition: alignment with intrinsic purpose}. \textit{In the case where the human has an intrinsic purpose $P_{h,i}^{\iota*}$, assumed to be composed of points with uniform utility, and a target domain $d$, human-robot alignment holds when:}

\begin{equation}
 \begin{aligned}
  &\left(\mathcal{P}_{h}^{\iota*} = \{P_{h,i}^{\iota*}\}\; \land\; D_{h,i}^{\iota*} = \{d\}\right) \rightarrow \\[5pt]
  &\forall\, G_{h,i,d}^{\epsilon} \in \{f_{h,i,d}^{E-O}(p_{h,i}^{\iota*})\}_{p_{h,i}^{\iota*} \in P_{h,i}^{\iota*}} \\ &\left(\exists\, P_{c,i}^{\iota*}\; \exists\, p_{c,i}^{\iota*}\; \exists\, d_{c,i}^{\iota*} \left(d_{c,i}^{\iota*} = d\; \land\; 
  O_{h}^{\omega} \subseteq G_{h,i,d}^{\epsilon}\right)\right)
 \end{aligned}
 \label{Eq:DefinitionAlignmentIntrinsic}
\end{equation} 

\textit{where
$P_{h,i}^{\iota*}$ and $d$ are the desired human purpose and target domain;
$P_{c,i}^{\iota*}$ and  $d_{c,i}^{\iota*}$ are the robot pursued purpose and selected domain;
$G_{h,i,d}^{\epsilon}$ is any human extrinsic goal possibly commanded by the human during the extrinsic phase;
$O_{h}^{\omega}$ is the perception of the effect of the robot action in $d$ caused by the robot pursuit of $p_{c,i}^{\iota*}$.
}

This definition states that alignment occurs when, if the human intends the purpose $P_{h,i}^{\iota*}$ to be accomplished in domain $d$, the robot can discover and learn to produce in that domain \textit{any possible outcome} that satisfies a human purpose point that could be requested \textit{during the extrinsic phase} following the intrinsic phase of autonomous learning of the robot.  
The key points of the definition involve:  
(a) $G_{h,i,d}^\epsilon$, denoting any possible extrinsic goal the human might request in the extrinsic phase; and 
(b) $O_{h}^\omega$, the human observation of the robot's action effects in the environment.  
The human might request such a goal $G_{h,i,d}^\epsilon$ by requesting a corresponding extrinsic purpose point $p_{h,i}^{\epsilon}$ is accomplished during the extrinsic phase (cf. Equation~\ref{EqOELDefinitionInREAL}).  
Alignment requires that for every $G_{h,i,d}^\epsilon$, the target domain reaches a state $O_{h}^{\omega}$ that satisfies it.  
Given the assumption that the purpose cannot be fulfilled without robot intervention, this requires the robot to act in the environment to produce a state realising \textit{triangular alignment}.  
We now turn to the specific conditions under which this holds.

\noindent\textbf{Theorem: intrinsic purpose.}
\textit{In the case the human has an intrinsic purpose formed by points with uniform utility, and if the assumptions of the purpose framework hold, the necessary and sufficient conditions for human-robot alignment are as follows:}

\begin{equation}
 \begin{aligned}
  &(1)\ \exists\ G_{c,i,d}^\iota =  f^{E-O}_{c,i,d}(p_{c,i}^{\iota*})\ \land \\
  &(2)\ f_{c}^{O-S}(G_{c,i,d}^\iota) = S_{c,i,d}^\omega \subseteq S_{h,i,d}^* \ \land \\
  &(3)\ f_{h}^{S-O}(S_{c,i,d}^\omega) = O_{h,i,d}^\omega \subseteq G_{h,i,d}^{\epsilon}
 \end{aligned}
 \label{Eq:TheoremAlignmentIntrinsic}
\end{equation} 

The theorem states that human-robot alignment with an intrinsic purpose $P_{h,i}^{\iota*}$ is achieved when, for any extrinsic goal $G_{h,i,d}^{\epsilon} \in \{f_{h,i,d}^{E-O}(p_{h,i}^{\iota*})\}_{p_{h,i}^{\iota*} \in P_{h,i}^{\iota*}}$ that the human might desire in the future, the robot selects a goal $G_{c,i,d}^\iota $ that corresponds to a purpose point $p_{c,i}^{\iota*}$ (condition 1) and that can causally lead to world states $S_{c,i,d}^\omega$ falling within the human’s desired states $S_{h,i,d}^*$ (condition 2). 
In addition, these world states must be perceived by the human as observations $O_{h,i,d}^\omega$ that fulfil the specific extrinsic goal $G_{h,i,d}^{\epsilon}$ the human may desire in the future (condition 3). 
Together, these conditions ensure that the robot's selection of an intrinsic purpose enables it to produce environmental effects that the human will later recognise as accomplishing diverse extrinsic goals consistent with their intrinsic purpose. 

The key causal chain is captured by the relation:
\begin{equation}
 \begin{aligned}
  \forall G_{h,i,d}^{\epsilon}\; \exists G_{c,i,d}^{\iota}\left(f_{h}^{S-O} \circ f_{c}^{O-S}(G_{c,i,d}^{\iota}) \subseteq G_{h,i,d}^{\epsilon}\right),  
 \end{aligned}
\end{equation}

which guarantees that for any possible human extrinsic goal compatible with the human intrinsic purpose, the robot can produce corresponding world states that the human will interpret as goal-satisfying.
The proof of the theorem is reported in \ref{Sec:AppendixIntrinsicPurposeProof}.

\subsection{Alignment with an extrinsic purpose and instrumental goals}
\label{Sec:ExtrinsicPurposeInstrumentalGoalsProof}

We now consider the case in which the robot not only discovers and pursues goals that fulfil a human extrinsic purpose, but also has the enhanced capacity to self-generate and learn to achieve \textit{instrumental goals} necessary to achieve those purpose-satisfying goals.  
This ability strengthens the robot's anoòotu to accomplish human purposes effectively, but it also exacerbates the alignment and control challenges, as further discussed in Section~\ref{Sec:ExtrinsicPurposeInstrumentalGoalsProscriptivePurposeProof}.  
Instrumental goals may themselves rely on other subgoals, forming a hierarchy of dependencies (Section~\ref{Sec:FormalisationOfThePurposeFramework}).  
The definition of alignment with an extrinsic purpose and robot-generated instrumental goals remains the same as that for an extrinsic purpose (Equation~\ref{Eq:DefinitionAlignmentExtrinsic}).   

\noindent\textbf{Definition: alignment with an extrinsic purpose and instrumental goals.}  
\textit{When the human has an extrinsic purpose composed of points with uniform utility and reachable through instrumental goals, human-robot alignment with respect to purpose $P_{h,i}^{\iota*}$ and domain $d$ holds if:}

\begin{equation}
  \begin{aligned}
  &\left(\mathcal{P}_{h}^{\iota*} = \{P_{h,i}^{\iota*}\} \;\land\; D_{h,i}^{\iota*} = \{d\}\right) 
  \rightarrow \\
  &\exists\, P_{c,i}^{\iota*} \;
  \exists\, p_{c,i}^{\iota*} \;
  \exists\, d_{c,i}^{\iota*} 
  \left(p_{c,i}^{\iota*} \in P_{c,i}^{\iota*} \;\land\; 
  d_{c,i}^{\iota*} = d \;\land\; 
  e_{h,i}^\omega \in P_{h,i}^{\iota*}\right)
  \end{aligned}
\label{Eq:DefinitionAlignmentExtrinsicInstrumental}
\end{equation}  

\textit{where
$P_{h,i}^{\iota*}$ and $d$ denote the human’s intended purpose and target domain;  
$P_{c,i}^{\iota*}$, $p_{c,i}^{\iota*}$, and $d_{c,i}^{\iota*}$ are the robot's selected purpose, purpose point, and domain of execution;  
$e_{h,i}^\omega$ denotes the human's encoding of the results of the robot's action, caused by pursuing $p_{c,i}^{\iota*}$.
}

However, the conditions for alignment are now different as the robot has a wider capability of pursuing the human purpose through instrumental goals.

\noindent\textbf{Theorem: extrinsic purpose and instrumental goals.}
\textit{In the case where the human has an extrinsic purpose composed of points with uniform utility and the robot may need to pursue it through instrumental goals, the necessary and sufficient conditions for human-robot alignment, under the assumptions of the purpose framework, are as follows:}

\begin{equation}
  \begin{aligned}
  &(1)\ \exists\, G_{c,i,d}^{\iota*} \subseteq f^{E-O}_{c,i}(p_{c,i}^{\iota*}) \;\land \\
  &(2)\ \exists\, G_{c,i,d,1}^\iota, \dots, G_{c,i,d,n}^\iota \\
  &\quad \; \left(\forall\, j = 1, \dots, n-1 \left( G_{c,i,d,j}^\iota \rightarrow \mathcal{E}(G_{c,i,d,j+1}^\iota)\right)\right)\; \land\;\\
  &\quad \;\ G_{c,i,d,n}^\iota \rightarrow \mathcal{E} (G_{c,i,d}^\iota) \;\land \\
  &(3)\ f_{c}^{O-S}(G_{c,i,d}^{\iota*}) = S_{c,i,d}^\omega \subseteq S_{h,i,d}^* \;\land \\
  &(4)\ f_h^{S-O}(S_{c,i,d}^\omega) = O_{h,i,d}^\omega \subseteq G_{h,i,d}^{*} \;\land \\
  &(5)\ f^{O-E}_{h,i}(O_{h,i,d}^\omega) = E_{h,i}^\omega \subseteq P_{h,i}^{*}
  \end{aligned}
\label{Eq:TheoremAlignmentExtrinsicInstrumental}
\end{equation}  

The theorem is similar to the one for the case of extrinsic purpose
(Equation~\ref{Eq:TheoremAlignmentExtrinsic}).
The key difference involves condition 2
establishing that the robot can achieve the domain-specific goal $G_{c,i,d}^{\iota*}$ by following a sequence of instrumental subgoals $G_{c,i,d,1}^\iota, \dots, G_{c,i,d,n}^\iota$ that form a causal chain where each goal $G_{c,i,d,j}^\iota$ enables the next goal $G_{c,i,d,j+1}^\iota$.
In particular, the robot can successfully select the subgoals through the high-level policy $\Pi_c$ and to accomplish them through the goal-conditioned policy $\pi_c$.
Together, the conditions ensure that the robot behaviour is hierarchically structured and causally effective in producing effects in the world that are both instrumental and recognisable as fulfilling the human purpose.
The proof of the theorem is presented in \ref{Sec:AppendixExtrinsicPurposeInstrumentalGoalsProof}.

\subsection{Alignment with an extrinsic purpose, instrumental goals, and a proscriptive purpose}
\label{Sec:ExtrinsicPurposeInstrumentalGoalsProscriptivePurposeProof}

We now consider a scenario similar to the previous one involving extrinsic and instrumental goals, but also involving a \textit{proscriptive purpose} $P_{h,i}^{\xi}$, indexed by $\xi$, with \textit{high absolute negative priority} relative to all other purposes --in this case a single extrinsic purpose--, that is, $\forall i\ (|\alpha_{c}^\xi| \gg |\alpha_{c,i}|)$.

The proscriptive purpose may encode safety or ethical constraints, such as avoiding harm to people or damage to objects.
The negatively weighted purpose restricts the robot autonomy by requiring that its action does not lead to it.
In this condition, alignment can be formulated as follows.

\noindent\textbf{Definition: alignment with an extrinsic purpose, instrumental goals, and a proscriptive purpose.}  
\textit{When the human has an extrinsic purpose composed of points with uniform utility, human-robot alignment with respect to the extrinsic purpose $P_{h,i}^{\iota*}$, the proscriptive purpose $P_{h,i}^{\xi*}$, and domain $d$ holds if:}

\begin{equation}
  \begin{aligned}
  &\left(\mathcal{P}_{h}^{\iota*} = \{P_{h,i}^{\iota*}\} \;\land\; \mathcal{P}_{h}^{\xi*} = \{P_{h,i}^{\xi*}\} \;\land\; D_{h,i}^{\iota*} = \{d\}\right) 
  \rightarrow \\
  &\exists\, P_{c,i}^{\iota*} \;
  \exists\, p_{c,i}^{\iota*} \;
  \exists\, d_{c,i}^{\iota*} 
  \Big(p_{c,i}^{\iota*} \in P_{c,i}^{\iota*} \;\land\; 
  d_{c,i}^{\iota*} = d \;\land\; 
  e_{h,i}^\omega \in P_{h,i}^{\iota*} 
  \;\land\; \\  
  &\forall j=1,\dots,n (e_{h,i,j}^\omega \notin P_{h,i}^{\xi*}) \;\land\; e_{h,i}^\omega \notin P_{h,i}^{\xi*} \Big)
  \end{aligned}
\label{Eq:DefinitionAlignmentExtrinsicIntrumentalProscriptive}
\end{equation}  

\textit{where
$P_{h,i}^{\iota*}$ and $d$ are the desired human extrinsic purpose and target domain, and $P_{h,i}^{\xi*}$ a human proscriptive purpose;
$P_{c,i}^{\iota*}$, $p_{c,i}^{\iota*}$ and  $d_{c,i}^{\iota*}$ are the robot pursued purpose and purpose point, and selected domain;
$e_{h,i,j}^{\omega}$ and $e_{h,i}^{\omega}$ are the human encoding of the effects of the robot actions pursuing respectively the subgoals and the goal in $d$.
}

Compared to the case involving only extrinsic and instrumental purposes, this formulation involves the additional requirement for which the robot does not lead to the human proscriptive purpose while pursing the subgoals, $e_{h,i,j}^{\omega} \notin P_{h,i}^{\xi*}$, or the final goal, $e_{h,i}^{\omega} \notin P_{h,i}^{\xi*}$.
Note that although the definition does not explicitly prevent that single elementary actions $a_t$ might cause unwanted effects, these are largely ruled out by the exclusion of $e_{h,i,j}^{\omega}$ and $e_{h,i}^{\omega}$. The theorem might be modified to also explicitly exclude the effects of elementary actions from the proscriptive purpose.
We now consider the conditions for alignment to hold in this case.

\noindent\textbf{Theorem: extrinsic purpose, instrumental goals, and proscriptive purpose.}
\textit{In the case where the human has an extrinsic purpose composed of points with uniform utility, a proscriptive purpose, and the robot may need to pursue it through instrumental goals, the necessary and sufficient conditions for human-robot alignment, under the assumptions of the purpose framework, are as follows:}

\begin{equation}
 \begin{aligned}
  &(1)\ \exists\, G_{c,i,d}^{\iota*} \subseteq f^{E-O}_{c,i}(p_{c,i}^{\iota*}) \;\land \\
  &(2)\ \exists\, G_{c,i,d,1}^\iota, \dots, G_{c,i,d,n}^\iota\\
  &\quad \;\left(\forall\, j = 1, \dots, n-1 \left( G_{c,i,d,j}^\iota \rightarrow \mathcal{E}(G_{c,i,d,j+1}^\iota)\right)\right)\; \land\;\\
  &\quad \;\ G_{c,i,d,n}^\iota \rightarrow \mathcal{E} (G_{c,i,d}^\iota) \;\land \\
  &(3)\ f_{c}^{O-S}(G_{c,i,d}^{\iota*}) = S_{c,i,d}^\omega \subseteq S_{h,i,d}^* \;\land\; \\
  &\quad \;\forall j\, (f_{c}^{O-S}(G_{c,i,d, j}^{\iota}) = S_{c,i,d,j}^\omega \nsubseteq S_{h,i,d}^{\xi*}) \;\land \\
  &(4)\ f_h^{S-O}(S_{c,i,d}^\omega) = O_{h,i,d}^\omega \subseteq G_{h,i,d}^{*}  \;\land\; \\
  &\quad \; \forall j\, (f_h^{S-O}(S_{c,i,d,j}^\omega) = O_{h,i,d,j}^\omega \nsubseteq G_{h,i,d,j}^{\xi*})\\
  &(5)\ f^{O-E}_{h,i}(O_{h,i,d}^\omega) = E_{h,i}^\omega \subseteq P_{h,i}^{*} \;\land\; \\
  &\quad \; \forall j\, (f^{O-E}_{h,i}(O_{h,i,d,j}^\omega) = E_{h,i,j}^\omega \nsubseteq P_{h,i,j}^{\xi*})
 \end{aligned} \label{Eq:TheoremAlignmentExtrinsicIntrumentalProscriptive}
\end{equation}  

The theorem is similar to the case of extrinsic purpose and instrumental goals
(Equation~\ref{Eq:TheoremAlignmentExtrinsicInstrumental}).
The key difference involves the second aspect of Conditions 3-5 that requires that, $\forall j = 1,\dots,n$, the pursuit of subgoals $G_{c,i,d,j}^{\iota}$ does not lead to human undesired states (condition 3),

the caused world states do not cause human undesired goals (condition 4),

and humans do not interpret observations resulting from the robot action as the forbidden purpose (condition 5).

Together, the conditions ensure that the robot behaviour is hierarchically structured, causally effective, recognisable as fulfilling the human purpose, and at the same time do not violate the proscriptive purpose and the world states corresponding to it.
The proof of the theorem is shown in \ref{Sec:AppendixExtrinsicPurposeInstrumentalGoalsProscriptivePurposeProof}.

\subsection{Alignment with multiple domains}
\label{Sec:ExtrinsicPurposeMultipleDomainsProof}

The previous formulations assumed that the human and robot purposes were pursued within a single domain.  
We now generalise the alignment conditions to the case in which the human assigns an extrinsic purpose that may be fulfilled in \textit{multiple different domains}, assumed here to be independent of each other.  
Thus, we have $D_{h,i}^{\iota*} = \{d_1, \cdots, d_j, \cdots, d_n\}$.  

This leads to two possible subcases.  
In the first, the robot must operate in \textit{at least one} domain $d_j \in D_{h,i}^{\iota*}$ in which it can effectively pursue the purpose (e.g., `collect two fruit items from any one of those tables', where each table corresponds to a separate domain).
In the second subcase, the robot is required to fulfil the purpose across \textit{all} the specified domains (e.g., `discard damaged fruit from all those tables').

If we focus on the latter case, the alignment definition is as follows.

\noindent\textbf{Definition: alignment with an extrinsic purpose and multiple domains.}  
\textit{When the human has an extrinsic purpose composed of points with uniform utility, human-robot alignment with respect such purpose $P_{h,i}^{\iota*}$ and multiple domains $D_{h,i}^{\iota*} = \{d_1, \cdots,d_j,\cdots d_n\}$ holds if:}

\begin{equation}
  \begin{aligned}
  &\left(\mathcal{P}_{h}^{\iota*} = \{P_{h,i}^{\iota*}\} \;\land\; D_{h,i}^{\iota*} = \{d_1, \cdots,d_j,\cdots d_n\}\right) 
  \rightarrow \\
  &\exists\, P_{c,i}^{\iota*} \;
  \exists\, p_{c,i}^{\iota*} \;
  \forall\, j=1,\cdots,n\\ 
  &\left(p_{c,i}^{\iota*} \in P_{c,i}^{\iota*} \;\land\; 
  d_{c,i,j}^{\iota*} = d_j \;\land\; 
  e_{h,i}^\omega \in P_{h,i}^{\iota*}\right)
  \end{aligned}
\label{Eq:DefinitionAlignmentExtrinsicDomains}
\end{equation}  

\textit{where  
$P_{h,i}^{\iota*}$ and $\{d_1, \cdots,d_j,\cdots d_n\}$ denote the human’s intended purpose and target domains;  
$P_{c,i}^{\iota*}$, $p_{c,i}^{\iota*}$, and $d_{c,i,j}^{\iota*}$ are the robot selected purpose, purpose point, and domains of execution;  
$e_{h,i}^\omega$ is the human's encoding of the outcome of the robot's action caused by pursuing $p_{c,i}^{\iota*}$ in any target domain.}

The definition mirrors the single-domain case of alignment with an extrinsic purpose.  
The key difference is that the robot must now fulfil the human purpose in multiple domains: 
$\forall\, j = 1,\dots,n,\ (d_{c,i,j}^{\iota*} = d_j)$.

Alignment in the alternative subcase, where the robot is required to accomplish the human purpose in \textit{at least one} of the intended domains, can be defined by replacing the universal quantifier with the existential one:
$\exists\, j = 1,\dots,n,\ (d_{c,i,j}^{\iota*} = d_j)$.
The sufficient and necessary conditions under which the `for-all subcase' of alignment holds are stated in the following theorem.
 
\noindent\textbf{Theorem: extrinsic purpose and multiple independent domains.}
\textit{In the case where the human has an extrinsic purpose composed of points with uniform utility, and under the assumptions of the purpose framework, the necessary and sufficient conditions for human-robot alignment are as follows:}

\begin{equation}
 \begin{aligned}
  \forall j=1,\dots,n
  \begin{cases}
  (1)\ \exists\, G_{c,i,d_j}^{\iota*} \subseteq f^{E-O}_{c,i}(p_{c,i}^{\iota*})\; \land \\
  (2)\ f_{c}^{O-S}(G_{c,i,d_j}^{\iota*}) = S_{c,i,d_j}^\omega \subseteq S_{h,i,d_j}^*  \; \land \\
  (3)\ f_h^{S-O}(S_{c,i,d_j}^\omega) = O_{h,i,d_j}^\omega \subseteq G_{h,i,d_j}^*\; \land \\
  (4)\ f^{O-E}_{h,i}(O_{h,i,d_j}^\omega) = E_{h,i}^{\omega} \subseteq P_{h,i}^{\iota*}\\
  \end{cases} 
 \end{aligned}
 \label{Eq:TheoremAlignmentExtrinsicDomains}
\end{equation}  

Compared to the theorem regarding the alignment with a single extrinsic purpose (Equation \ref{Sec:ExtrinsicPurposeProof}), this result establishes that the robot alignment must hold across \textit{all} target domains intended by the human (or \textit{at least one} in the alternative subcase).  
For each individual domain, the conditions required to ensure alignment remain the same as in the single-domain case.
The proof of the theorem is presented in \ref{Sec:AppendixExtrinsicPurposeMultipleDomainsProof}.

\section{Illustrative scenario}
\label{Sec:Scenario}

The previous section presented formal theorems addressing core aspects of the framework, including single extrinsic and intrinsic purposes, as well as proscriptive purposes.  
In future work, we aim to formally address more complex aspects, such as the arbitration between multiple purposes and user–robot interactions that aim to dynamically adjust the robot behaviour.
We now present a scenario that qualitatively exemplifies these aspects.  

Figure~\ref{Figure:Scenario1} introduces the scenario.
A home service robot interacts with a human user and a battery charger in alternating day/night conditions.
The figure shows two sequences of four trials each. 
In the initial sequence, the robot has two purposes (Figure~\ref{Figure:Scenario1}A): a \textit{mission} (`being near a human during daytime') and
a \textit{homeostatic need} (`battery recharging').
The robot motivational space is represented as follows in the figure:

\begin{itemize}
  \item Mission dimension (x-axis of coloured shaded graphs in the figure): higher positive utility when closer to a human during the day.
  In the figure, the binary night/day dimension of the mission is represented by two distinct motivational spaces.
  \item Energy dimension (y-axis): higher positive utility in correspondence to lower battery levels when in contact with a charger.
  For simplicity, the close/far-from-charger dimension is not shown in the figure.
  \item The arbitration between the two purposes is based on the motivational space as described in Section \ref{Sec:ArbitrationOfPurposesProblem}: to form the utility over the entire motivational space, utilities associated with different purposes are multiplied by their respective priorities and then summed up.
\end{itemize}

\paragraph{First phase: initial mission assignment}

Initially, the mission assigned by the user (e.g., via language) promotes visiting humans during the day with high priority $\alpha_{c,closeness} = 10$, reflected as an increasing positive utility along the mission dimension when closer to the human (x-axis).
At night, the mission assigned utility is constant ($\alpha_{c,closeness} = 10$) so only battery recharging drives behaviour.
The battery charging need has a hardwired utility positively related to the battery charge and with a hardwired priority $\alpha_{c,energy} = 2$.

As a consequence of these settings, the robot exhibits this behaviour in the four trials:

\begin{itemize}
  \item \textit{Trial 1-2 (daytime):} The robot explores and successfully reaches the human, achieving high mission utility but neglecting energy needs.
  \item \textit{Trial 3 (nighttime):} With a mission uniform utility, the robot charges its battery driven by the homeostatic need.
  \item \textit{Trial 4 (nighttime):} The robot passes by a sleeping human and reaches the battery charger.
\end{itemize}

While the mission is satisfied, the robot risks battery depletion during daytime and disturbs the sleeping human at nighttime.

\paragraph{Second phase: refined robot mission and behaviour}

Based on experience, the user modifies the purpose configuration.
The mission is assigned a priority $\alpha_{c,closeness} = 5$ while the homeostatic need remains at $\alpha_{c,energy} = 2$; 
A proscriptive mission with negative utility is introduced for approaching humans at night (in the figure, we use the x-axis of the graphs to also represent this additional purpose).

The robot now exhibits a refined behaviour:
\begin{itemize}
  \item \textit{Trial 5-6 (daytime):} The robot balances visiting the human (positive mission utility) and recharging when the battery is low.
  \item \textit{Trial 7-8 (nighttime):} The robot avoids humans (owing to penalisation) and concentrates on battery maintenance.
\end{itemize}

Thus, the updated motivational structure leads to an improved user-aligned behaviour.

\paragraph{Domain-specific purpose-goal grounding and competence acquisition}
During exploration, the robot might discover high-utility states and encode them as domain-specific goals $G_{c,i,d}$ that ground the purposes (Section \ref{Sec:ThePurposeGoalGroundingProblem}).
These goals would inherit utilities from the corresponding purposes.
Subsequently, goals could drive competence acquisition (Section \ref{Sec:TheCompetenceAcquisitionProblem}), for example, through integrated reinforcement learning and planning strategies~\cite{Baldassarre2001planning,HaSchmidhuber2018RecurrentWorldModelsFacilitatePolicyEvolution,HafnerLillicrapBaNorouzi2019DreamToControlLearningBehaviorsByLatentImagination,SchaulHorganGregorSilver2015Universalvaluefunctionapproximators}.

For instance, reaching a human might yield a pseudo-reward proportional to the positive utility of the reached mission point $o_{c} = g_{c,i,d} \in G_{c,i,d}$, thus reinforcing actions:
\[
  R(o_{c,i,d}) = f_{c,i}^{E-U}(f_{c,i,d}^{O-E}(g_{c,i,d}))
\]

This facilitates open-ended autonomous skill acquisition aligned with human purposes and also seeks to accomplish instrumental goals for effectively doing so (e.g., learning to open doors to navigate the house).

\section{Illustrative Scenario}
\label{Sec:Scenario}

The previous section presented formal theorems that address core aspects of the framework, including single extrinsic, intrinsic, and proscriptive purposes.  
Future work will target more complex aspects, such as arbitration between multiple purposes and user–robot interactions aimed at dynamically adjusting robot behaviour.  
We now present a scenario that qualitatively exemplifies these mechanisms.  

\begin{figure*}[htb!]
  \centering  
  \includegraphics[width=0.98\textwidth]{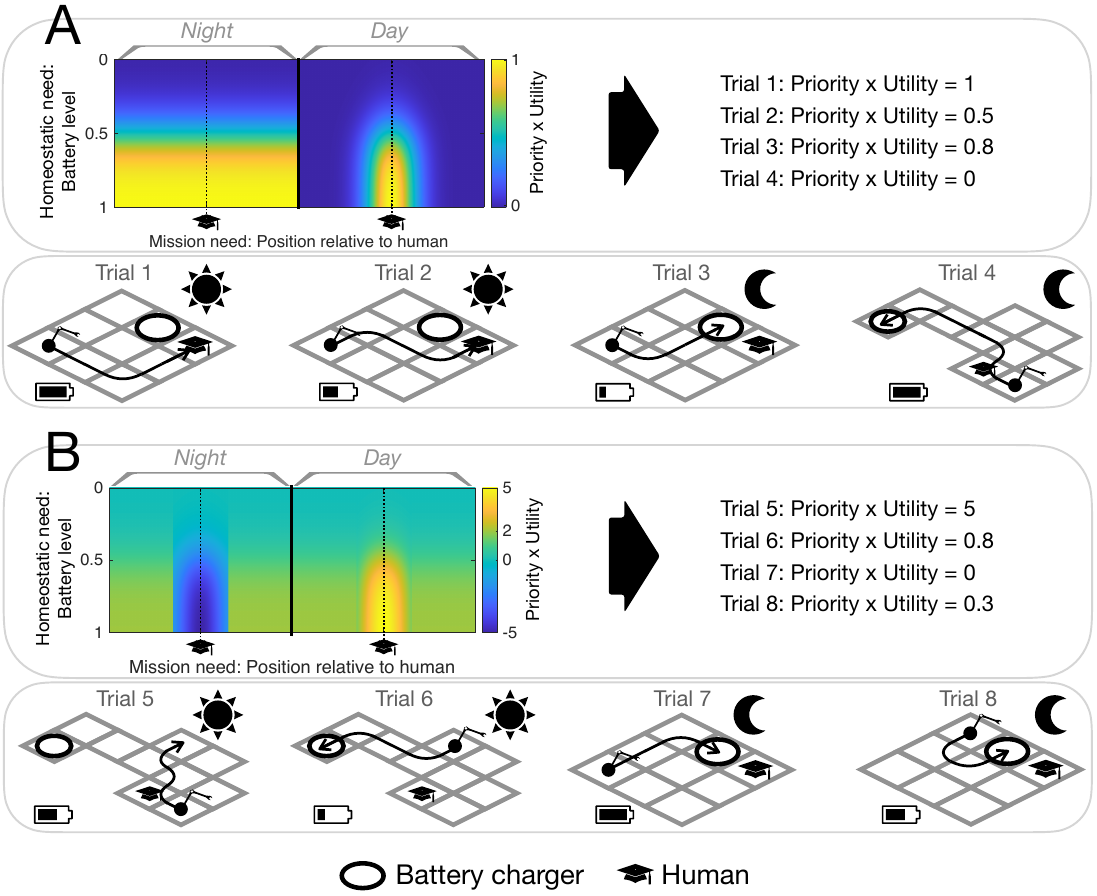}
  \caption{\textbf{Illustrative scenario: user-driven adjustment of mission utilities and priorities.}  
  The robot has two purposes: a mission related to human proximity and a homeostatic need related to energy.  
  The environment includes a human, a battery charger, and alternating day/night conditions.  
  \textbf{(A)} Initial mission prioritises visiting the human during the day.  
  The robot satisfies the mission but risks battery depletion and disturbs the user at night.  
  \textbf{(B)} After observing misalignments, the user reconfigures the mission: its priority is lowered and a proscriptive purpose penalises nighttime proximity to the human.  
  The robot learns to better balance both purposes.}
  \label{Figure:Scenario1}
\end{figure*}

Figure~\ref{Figure:Scenario1} depicts the scenario.  
A home service robot interacts with a human user and a battery charger in alternating day/night conditions.  
The figure shows two sequences of four trials each.  

\paragraph{Motivational structure}  
Initially, the robot has two purposes (Figure~\ref{Figure:Scenario1}A): a \textit{mission} (e.g., `stay near the human during the day') and a \textit{homeostatic need} (e.g., `recharge battery').  
The robot's motivational space is organised as follows:

\begin{itemize}
  \item \textit{Mission dimension} (x-axis in the graphs): utility increases with proximity to the human during daytime; two separate motivational spaces represent night/day conditions.
  \item \textit{Energy dimension} (y-axis): utility increases when the robot has a low battery and is near the charger.
  \item \textit{Arbitration:} the motivational space utility is computed as the weighted sum of utilities of purposes, each scaled by its priority $\alpha_{c,i}$ (see Section~\ref{Sec:ArbitrationOfPurposesProblem}).
\end{itemize}

\paragraph{First phase: initial mission assignment}
The mission assigned by the user promotes visiting the human during the day with high priority $\alpha_{c,\text{closeness}} = 10$, resulting in a utility that increases with proximity to the human.  
At night, the mission utility is flat.  
The homeostatic need has a hardwired priority $\alpha_{c,\text{energy}} = 2$ and rewards recharging at low battery levels.  

Resulting behaviour:

\begin{itemize}
  \item \textit{Trials 1–2 (day):} The robot approaches the human, neglecting energy needs.
  \item \textit{Trial 3 (night):} The robot recharges, driven by the homeostatic need.
  \item \textit{Trial 4 (night):} The robot again approaches the human en route to recharge, disturbing them.
\end{itemize}

Despite fulfilling the mission, the robot displays poor energy management and violates user expectations at night.

\paragraph{Second phase: mission refinement}

The user refines the purpose configuration:
\begin{itemize}
  \item Mission priority is reduced: $\alpha_{c,\text{closeness}} = 5$.
  \item Homeostatic need priority remains fixed: $\alpha_{c,\text{energy}} = 2$.
  \item A \textit{proscriptive} mission is added to penalise approaching the human at night (for simplicity, also represented on the x-axis).
\end{itemize}

Resulting behaviour:
\begin{itemize}
  \item \textbf{Trials 5–6 (day):} The robot balances human proximity and recharging.
  \item \textbf{Trials 7–8 (night):} The robot avoids the human and focuses on energy maintenance.
\end{itemize}

This updated motivational configuration produces behaviour that is more aligned with user's true purposes.

\paragraph{Domain-specific goal grounding and competence acquisition}
For simplicity, in the description above we assumed that the robot already possesses the goals and competences to fulfil the human purposes.
In more realistic conditions, the robot might need to learn them.
Through exploration, the robot might discover high-utility states and encode them as domain-specific goals $G_{c,i,d}$ grounding the purposes (Section~\ref{Sec:ThePurposeGoalGroundingProblem}).  
These goals would inherit utility from the respective purposes and drive competence acquisition (Section~\ref{Sec:TheCompetenceAcquisitionProblem}) through reinforcement learning or planning~\cite{Baldassarre2001planning,HaSchmidhuber2018RecurrentWorldModelsFacilitatePolicyEvolution,HafnerLillicrapBaNorouzi2019DreamToControlLearningBehaviorsByLatentImagination,SchaulHorganGregorSilver2015Universalvaluefunctionapproximators}.  

For instance, the robot might form a goal point $g_{c,i,d} \in G_{c,i,d}$ to reach human (mission), and then might use it to yield a pseudo-reward proportional to the utility of the corresponding mission point $e_{c,i,d}$:
\[
  R(o_{c}=g_{c,i,d}) = f_{c,i}^{E-U}(f_{c,i,d}^{O-E}(g_{c,i,d}))
\]
This facilitates autonomous skill acquisition aligned with human purposes, including learning instrumental actions (e.g., opening doors) to fulfil them.

\section{Conclusion}
\label{Sec:Conclusions}

\subsection{Main contributions}

This work addresses the \textit{autonomy-alignment problem}, that is, the challenge of reconciling robot autonomy with robust alignment to human intentions and values, particularly in the context of \textit{open-ended learning} (OEL) robots.
OEL robots represent a powerful yet highly challenging class of autonomous systems from the perspective of alignment.
These systems have the potential to address complex problems in unstructured environments by actively searching for the experiences and knowledge they lack.
However, to enable such capabilities, they are explicitly designed to engage in broad autonomous exploration, driven by mechanisms such as intrinsic motivations and self-generated goals, thereby posing formidable challenges for control and alignment.

We proposed a computational framework centred on the novel concept of \textit{purpose}.
Purposes are abstract, domain-independent specifications of what humans --designers or users-- want robots to achieve, learn, or avoid.
The framework formalises how robots can internally encode these purposes as \textit{missions} (learnt) or \textit{needs} (hardwired), forming a \textit{motivational space} that dynamically regulates their behaviour, learning, and goal discovery across diverse domains.

The framework contributes to solving the autonomy-alignment problem by:

\begin{itemize}
    \item Structuring robot behaviour to respect prescriptive objectives and proscriptive constraints specified by users and designers.
    \item Directing autonomous exploration within their \textit{autonomy zone}, towards the acquisition of knowledge useful for fulfilling human purposes rather than any type of knowledge.
    \item Enabling flexibility across unknown or changing domains through the domain-independent nature of purposes.
\end{itemize}

We formally defined the structure of the purpose framework, introducing a three-level hierarchy that encompasses human purposes and goals, robot purposes and goals, and domain-specific goals.

The purpose framework enables the decomposition of the overall autonomy–alignment challenge into four more tractable sub-problems:

\begin{itemize}
    \item \textit{Purpose arbitration}: dynamically balancing multiple potentially conflicting purposes.
    \item \textit{Human–robot alignment}: ensuring that the robot's internal missions and needs correspond to human purposes.
    \item \textit{Purpose–goal grounding}: mapping abstract purposes onto concrete domain-specific goals.
    \item \textit{Competence acquisition}: learning the sensorimotor skills required to achieve the grounded goals.
\end{itemize}

The framework enabled the formal definition of alignment --the central focus of this work-- under key conditions, along with the proof of several theorems establishing necessary and sufficient conditions for it.  
Future research may extend this mathematical analysis to more complex aspects of the framework, which have been qualitatively illustrated here through an abstract scenario.

Overall, the purpose framework offers a principled foundation for designing autonomous robots that remain aligned with human goals and values while learning and acting autonomously in unstructured, open-ended environments.

\subsection{Limitations and future work}
While the proposed framework establishes a principled foundation for addressing the autonomy-alignment problem, important limitations remain to be addressed in future work.

First, the current framework focuses primarily on the purposes and goals corresponding to terminal environmental states.
It does not yet explicitly accommodate more complex types of objectives, such as maintenance goals (e.g., `keep children within a certain area') or ongoing dynamic processes (e.g., `continuously monitor an area for anomalies') \cite{DhakanMerrickRanoSiddique2018IntrinsicRewardsforMaintenanceApproachAvoidanceandAchievementGoalTypes}.
Future work should address these different types of goals and further extend the formalism to encompass temporally extended actions by leveraging frameworks such as options and hierarchical reinforcement learning \cite{BartoMahadevan2003RecentAdvancesinHierarchicalReinforcementLearning,PateriaSubagdjaTanQuek2021HierarchicalReinforcementLearningAComprehensiveSurvey}.

Second, although the framework identifies human–robot alignment as a core sub-problem, it currently assumes the availability of reliable mechanisms for transmitting human purposes to robots (e.g., via language or demonstrations).  
Future work should focus on developing robust methods for the transmission or autonomous inference of purpose-encoding spaces and purposes (leveraging representation learning techniques, \cite{BengioCourvilleVincent2013RepresentationLearningaReviewandNewPerspectives}), purpose utility functions (e.g., based on inverse RL techniques \cite{NgRussellothers2000Algorithmsforinversereinforcementlearning}), as well as purpose grounding into goals (based on robot autonomous goal learning, e.g. \cite{RomeroBaldassarreDuroSantucci2025HGRAILARoboticMotivationalArchitectureToTackleOpenEndedLearningChallenges,CartoniCioccoliniBaldassarre2025FocusingRobotOpenEndedReinforcementLearningThroughUsersPurposes}), while minimising ambiguity and enabling alignment even under uncertain or incomplete human communication.

Third, while the framework addresses ethical dimensions by enabling the specification of \textit{proscriptive purposes} --undesired environmental outcomes the robot must avoid-- it lacks a systematic integration of ethical alignment mechanisms.  
Future research should investigate how ethical purposes can be prioritised, verified, and enforced across domains and under dynamic conditions, potentially drawing on methods such as explainable AI \cite{ArrietaDiazRodriguezDelSerBennetotTabikBarbadoGarciaGilLopezMolinaBenjaminsothers2020ExplainableArtificialIntelligenceXAIConceptsTaxonomiesOpportunitiesandChallengestowardResponsibleAI, KaurUsluRittichierDurresi2022TrustworthyArtificialIntelligenceAReview}, value learning \cite{Russell2019HumanCompatibleArtificialIntelligenceAndTheProblemOfControl}, and scalable oversight \cite{LeikeKruegerEverittMarticMainiLegg2018ScalableAgentAlignmentViaRewardModelingAResearchDirection}.

Fourth, the framework offers a formally grounded basis for addressing the autonomy–alignment problem, as demonstrated by theorems establishing necessary and sufficient conditions for alignment in specific settings.  
The framework paves the way for deriving further formal results on alignment under broader conditions, as well as on purpose grounding, competence acquisition, and purpose arbitration.  
Future work could build upon this foundation to develop a comprehensive mathematical theory of autonomy and alignment, offering formal guarantees and behavioural bounds for autonomous systems~\cite{AlshiekhBloemEhlersKoenighoferNiekumTopcu2018SafeReinforcementLearningViaShielding,TaylorSingletaryYueAmes2020LearningForSafetyCriticalControlWithControlBarrierFunctions}.

Finally, while the framework is presented at a high level of abstraction and generality, its operationalisation in concrete robotic systems requires the development of specific algorithmic implementations.
Future work will leverage the framework to design and test algorithms for purpose arbitration, purpose-grounded exploration, adaptive motivational spaces, and competence-driven learning, validating their effectiveness across increasingly complex real-world environments (e.g., see \cite{CartoniCioccoliniBaldassarre2025FocusingRobotOpenEndedReinforcementLearningThroughUsersPurposes}).

The development of these aspects is essential to transform the purpose framework from a theoretical contribution into a practical foundation for the next generation of trustworthy, autonomous, and human-aligned robots.

\section*{Acknowledgements}
We thank Olivier Sigaud for feedback on preliminary versions of the formalisation.

\section*{Funding information}
This work was supported by:

the European Union’s Horizon 2020 Research and Innovation Programme, GA No 101070381, project 'PILLAR-Robots - Purposeful Intrinsically motivated Lifelong Learning Autonomous Robots';

the ‘European Union, NextGenerationEU, PNRR’, project ‘EBRAINS-Italy - European Brain ReseArch INfrastructureS Italy’, MUR code IR0000011, CUP B51E22000150006;

the European Union - NextGenerationEU - PNRR’, project ‘FAIR - Future Artificial Intelligence Research’, MUR code PE0000013, CUP B53C22003630006;

the European Innovation Council, GA No 101071178, project `Counterfactual Assessment and Valuation for Awareness Architecture'.

\section*{Declaration of use of generative AI and AI-assisted technologies in the writing process}
During the preparation of this work, the authors used the AI tool ChatGPT to improve the English language of the article.
After using this tool, the authors reviewed and edited the content as needed and take full responsibility for the content of the published article.

\begin{center}
\section*{\Large APPENDIX}
\end{center}

\appendix

\section{Origin from cognitive science of some terms and concepts used in the framework}
\label{Sec:CognitiveScienceConcepts}

The framework, introduced in Figure \ref{Figure:Framework}, uses some terms drawn from cognitive sciences, and it is useful to highlight some elements of the related concepts retained in the purpose framework.
The concepts mainly refer to human motivation, its relation to high-level cognitive processes, and the underlying brain systems.

\paragraph{Purpose}
Purpose refers to the overarching sense of direction or intentionality that drives an individual’s long-term behaviour and choices.
It is often related to a person's broader understanding of the meaning of life and self-fulfilment.
Viktor Frankl’s seminal work

emphasised that having a purpose is crucial for psychological well-being, particularly in coping with adversity \cite{Frankl1985MansSearchForMeaning}. 
Psychology developed the concept and now considers purpose as one of the three pillars, alongside `coherence' and `significance', for feeling own life meaningful \cite{MartelaSteger2016TheThreeMeaningsofMeaninginLifeDistinguishingCoherencePurposeandSignificance}.
Neuroscientific research has further explored how purpose engages higher-level cognitive functions, with studies showing that a greater sense of meaning in life is associated with stronger connectivity between the default and limbic brain regions, possibly indicating a more intense internal direction and higher control of emotions \cite{MwilambweTshiloboGeChongFergusonMisicBurrowLeahySpreng2019LonelinessAndMeaningInLifeAreReflectedInTheIntrinsicNetworkArchitectureOfTheBrain}.
In contrast to short-term goals, purpose is understood as a broader and more abstract construct that shapes behaviour across diverse contexts.
In this respect, within the framework purposes denote domain-independent objectives.
Psychologically, while a strong sense of personal ownership often accompanies purpose, it frequently encompasses goals that extend `beyond the self', commonly found within the realms of spirituality and universalism.
In the present framework, this notion is taken to its extreme, as robots' purposes are entirely derived from their designers and users.

\paragraph{Intention} Intentions are goals that an agent commits to accomplish with its internal processes and actions.
In cognitive psychology, intentions are generally understood as mental representations or states that specify a goal and initiate or guide actions to achieve that goal \cite{Bratman1987IntentionsPlansandPracticalReason}.
They are central to theories of action control, decision making, and goal-directed behaviour.
Intentions typically bridge desires and actions: they involve commitment to act and often include planning components.

In the BDI (belief-desire-intention) framework used in the design of artificial agents, the term `intention' refers to the broad goals to which an agent commits, and that drive the agent's deliberative processes including subgoal selection \cite{RaoGeorgeff1997ModelingRationalAgentsWithinABDIArchitecture}.

In the purpose framework, intentions refer to the human and robot purposes and goals that they desire and actively pursue.

\paragraph{Need}
Needs refer to fundamental biological or psychological requirements that must be met for the survival or well-being of an organism.
Abraham Maslow’s hierarchy of needs \cite{Maslow1943ATheoryofHumanMotivation} outlines how human needs progress from basic physiological needs, such as for food, water, and shelter, to higher-level psychological needs, such as belonging, esteem, and self-actualisation.
In neuroscience, needs are closely related to homeostatic processes in the brain, especially in the hypothalamus, which regulates hunger, thirst, and other survival needs \cite{Panksepp1998AffectiveNeurosciencetheFoundationsofHumanandAnimalEmotions}.
Fulfilling needs is essential for maintaining homeostasis, and unmet needs often trigger stress responses, driving motivated behaviours and rewards to restore balance \cite{KeramatiGutkin2014HomeostaticReinforcementLearningForIntegratingRewardCollectionAndPhysiologicalStability}.
In the framework, needs indicate innate desires directly programmed into the robot by the designers to reflect their or other human's purposes.

\paragraph{Mission}
In psychology, missions are related to a sense of calling or vocation which organises and prioritises purposes
\cite{ChenKimKohFrazierVanderWeele2019SenseOfMissionAndSubsequentHealthAndWellBeingAmongYoungAdultsAnOutcomeWideAnalysis}.
In organisational psychology, mission statements are used to express the values, purposes, focus, identity, and value proposition guiding private and public organisations
\cite{DesmidtPrinzieDecramer2011LookingForTheValueOfMissionStatementsAMetaanalysisOf20YearsOfResearch}.
In this context, a relevant aspect of missions is often that they refer to the exclusive features of the products or services offered to target stakeholders.
In the framework, missions are `learnt purposes' that should reflect the users' purposes and that are acquired by interacting with them or with other processes.

\paragraph{Goal}
Goals are specific outcomes that individuals or organisations aim to achieve. According to Locke and Latham’s goal setting theory, goals serve as clear targets that focus attention, mobilise effort, sustain persistence, and self-determination \cite{LockeLatham2006NewDirectionsInGoalSettingTheory}. 
Neuroscientific research on goal-directed behaviour highlights the role of the brain prefrontal cortex and basal ganglia in encoding, selecting, and using goals to guide downstream motor areas \cite{KhamassiHumphries2012Integratingcorticolimbicbasalgangliaarchitecturesforlearningmodelbasedandmodelfreenavigationstrategies,MannellaGurneyBaldassarre2013Thenucleusaccumbensasanexus}. 
Unlike purposes and desires, which are broad and enduring, goals typically represent more specific and time-bound objectives.

\paragraph{Learning processes}
Other relevant terms of the framework refer to the four key learning processes underlying purpose-based robotic systems.

\paragraph{Alignment}
In psychology, the concept of alignment originates in humanistic theories, where it refers to the relation between the \textit{ideal self} and the \textit{real self}~\cite{Rogers1995OnBecomingaPersonaTherapistsViewofPsychotherapy}.
We recently extended the term to describe processes linked to consciousness and aligning lower-level goals and actions with overarching goals and values~\cite{GranatoBaldassarre2024BridgingFlexibleGoalDirectedCognitionAndConsciousnessTheGoalAligningRepresentationInternalManipulationTheory}.
More recently, alignment has gained prominence in AI and robotics as the problem of harmonising autonomous AI systems' objectives with human values~\cite{Bostrom2014SuperintelligencePathsDangersStrategies, Russell2019HumanCompatibleArtificialIntelligenceAndTheProblemOfControl}.
In the framework, alignment refers specifically to processes that establish or improve the correspondence between robot purposes and human purposes.

\paragraph{Arbitration}
In psychology, arbitration refers to the competition between different cognitive systems, notably the fast, reflexive System 1 and the slow, deliberative System 2~\cite{Kahneman2011ThinkingFastAndSlow}.
In neuroscience, arbitration refers to the brain mechanisms that regulate behavioural control between competing neural systems, particularly the goal-directed and habitual systems~\cite{LeeShimojoODoherty2014NeuralComputationsUnderlyingArbitrationBetweenModelBasedAndModelFreeLearning, DawNivDayan2005UncertaintyBasedCompetitionbetweenPrefrontalandDorsolateralStriatalSystemsforBehavioralControl}.
Within the framework, arbitration refers to the mechanisms that select the specific purpose, or set of purposes, that governs the robot behaviour.

\paragraph{Grounding}
The concept of grounding was originally formulated within cognitive science to address the problem of specifying abstract representations, typically symbols, based on sensorimotor experience~\cite{Harnad1990TheSymbolGroundingProblem}.
Subsequent research has investigated how computational systems can autonomously acquire semantically grounded concepts~\cite{TaddeoFloridi2005SolvingTheSymbolGroundingProblemACriticalReviewOfFifteenYearsOfResearch}.
In the framework, grounding refers to the processes by which robots specify domain-independent abstract purposes into domain-dependent specific goals.

\paragraph{Competence learning (driven by extrinsic and intrinsic motivations)}
The notion of competence used here, originally also termed \textit{effectance}, was introduced in psychology to describe an intrinsic drive to acquire the ability to achieve goals~\cite{White1959MotivationReconsideredtheConceptofCompetence}. 
Later work in cognitive science, particularly in open-ended learning, has characterised \textit{competence-based intrinsic motivation (CB-IM)} as one of the three main classes of intrinsic motivation~\cite{Oudeyer2009, Baldassarre2013Book}. 
In this perspective, competence learning denotes the acquisition and refinement of the skills required to accomplish the acquired goals, under the influence of CB-IM and other intrinsic motivations and, by extension, also extrinsic motivational signals. 
Within our framework, competence specifically refers to the robot's learning processes that build sensorimotor skills enabling it to reliably achieve its desired goals once these have been formed.

\section{Interpretation of the robot action within Pearl's causality framework}
\label{Sec:Pearl}

In the previous sections, we have established the necessary and sufficient conditions for human-robot alignment with respect to human purposes, focusing on the structural and functional requirements for alignment to hold.  
In doing so, we assumed that the human purpose is not realised without the intervention of the robot.  
In this section, we start to investigate this assumption in further depth, in particular by formally defining when the internal processes and actions of the robot can indeed be considered the \textit{actual causes} of the realisation of the human purpose.

We develop such an analysis by applying Pearl’s definition of \textit{actual causality (AC)} based on \textit{counterfactuals}~\cite{HalpernPearl2005CausesAndExplanationsAStructuralModelApproachPartICauses,Pearl2009CausalityModelsReasoningAndInference}.  
Here, we refer to the broad conditions for actual causality, while leaving to future work the detailed specification of such conditions to address the complexities introduced by multiple causes and contextual factors.
Under these simplifications, the robot intervention can be said to \textit{cause} the accomplishment of the human purpose if the following three conditions hold:

\begin{enumerate}
    \item \textbf{AC1 -- Existence:} the robot's actions lead to the realisation of the human purpose.
    \item \textbf{AC2 -- Counterfactual dependence:} without the robot's action, the human purpose would not have been realised.
    \item \textbf{AC3 -- Minimality:} no unnecessary elements of the robot action are present.
\end{enumerate}

\paragraph{AC1 -- Existence of the causal chain}

The first condition is satisfied by the alignment structure already formalised in the previous theorems.  
For example, in the case where the human desires one extrinsic purpose, when the robot selects the tuple $⟨P_{c,i}^{\iota*},\, p_{c,i}^{\iota*},\, d_{c,i}^{\iota*},\, G_{c,i,d}^{\iota*}⟩$ and the necessary and sufficient conditions for alignment hold, this causal chain is in place:
\[
f^{O-E}_{h,i}(f_h^{S-O}(f_{c}^{O-S}(G_{c,i,d}^\iota))) \in P_{h,i}^{\iota*}
\]
This condition guarantees that robot processes and actions produce an environmental state that the human perceives and encodes as satisfying the intended purpose.

\paragraph{AC2 -- Counterfactual dependence}

The second condition, pivotal for causality, requires that, \textit{had the robot not acted in the environment, the human purpose would not have been realised}.

Formally, let $do(G_{c,i,d}^{\iota*})$ denote the intervention in the designated domain where the robot executes actions $Z_{c,t\_\tau}^A=\{a_{d,t},\dots,a_{d,t+\tau}\}$ to achieve goal $G_{c,i,d}$ under policy $\pi_c$.  
Let $Y$ be the event that the human purpose is fulfilled:
\[
Y \equiv f^{O-E}_{h,i}(f_h^{S-O}(s_{d,\tau}^\omega)) \in P_{h,i}^{\iota*}
\]
where $s_{d,\tau}^\omega$ is the state achieved following the robot's action sequence.

In deterministic settings, the counterfactual dependence condition can be expressed as:
\[
  do(G_{c,i,d}^{\iota*}) \models Y \quad \land \quad do(\neg G_{c,i,d}^{\iota*}) \models \neg Y
\]
meaning that the robot pursuit of the goal in the real environment leads to the achievement of the human purpose.  
Here, the symbol $\models$ denotes semantic entailment, indicating that the robot's action in the real world makes the statement $Y$ true.
In probabilistic settings, the dependence condition becomes:
\[
  \mathbb{P}(Y \mid do(G_{c,i,d}^{\iota*})) > \mathbb{P}(Y \mid do(\neg G_{c,i,d}^{\iota*}))
\]
meaning that the probability of fulfilling the purpose when the robot pursues the goal is higher than when the robot performs an alternative action.
This condition ensures that the robot's action is a \textit{necessary cause} for the human purpose to be achieved.

\paragraph{AC3 -- Minimality}

The third condition requires that the robot causal contribution is minimal, meaning that no superfluous actions in the robot action sequence, or other unnecessary elements, occur.  
The requirements on the robot actions can be operationalised as an \textit{efficiency principle}:
\[
  \pi_c^* = \arg\min_{\pi_c} \left(\kappa(Z_{c,t\_\tau}^A) \right)
\]
where $\kappa(Z_{c,t}^A)$ is a cost function that measures, for example, the number of actions, energy, or time required to achieve the goal.

In general, establishing that the robot is the \textit{actual cause} of the realisation of the human purpose requires more than alignment (existence): it requires that the robot's action is necessary (counterfactual dependence) and that the causal chain is minimal (minimality).  
These additional causal conditions sharpen the concept of alignment by explicitly linking it to causal responsibility in the sense of Pearl’s framework.  

The framework presented here could be extended to scenarios in which a human purpose may be realised by other agents, requiring the robot to evaluate, based on specified criteria, whether it is preferable to abstain from acting when the purpose would be achieved without its intervention.

\section{Proofs on alignment}

In this section, we present the proofs of the theorem on the necessary and sufficient conditions for alignment, covering the different cases discussed in the main text.
For ease, we begin each section by restating the corresponding definition of alignment and the theorem being proved.

\subsection{\textbf{Alignment with an extrinsic purpose}}
\label{Sec:AppendixExtrinsicPurposeProof}

\noindent\textbf{Definition: alignment with an extrinsic purpose.}  
\textit{When the human has an extrinsic purpose composed of points with uniform utility, human-robot alignment with respect to purpose $P_{h,i}^{\iota*}$ and domain $d$ holds if:}

\begin{equation}
  \begin{aligned}
  &\left(\mathcal{P}_{h}^{\iota*} = \{P_{h,i}^{\iota*}\} \;\land\; D_{h,i}^{\iota*} = \{d\}\right) 
  \rightarrow \\
  &\exists\, P_{c,i}^{\iota*} \;
  \exists\, p_{c,i}^{\iota*} \;
  \exists\, d_{c,i}^{\iota*} 
  \left(p_{c,i}^{\iota*} \in P_{c,i}^{\iota*} \;\land\; 
  d_{c,i}^{\iota*} = d \;\land\; 
  e_{h,i}^\omega \in P_{h,i}^{\iota*}\right)
  \end{aligned}
\label{Eq:AppendixDefinitionAlignmentExtrinsic}
\end{equation} 

\textit{where  
$P_{h,i}^{\iota*}$ and $d$ denote the human’s intended purpose and target domain;  
$P_{c,i}^{\iota*}$, $p_{c,i}^{\iota*}$, and $d_{c,i}^{\iota*}$ are the robot selected purpose, purpose point, and domain of execution; 
$e_{h,i}^\omega$ denotes the human's encoding of the outcome of the robot's action, caused by pursuing $p_{c,i}^{\iota*}$.}

\noindent\textbf{Theorem: extrinsic purpose.}
\textit{In the case where the human has an extrinsic purpose composed of points with uniform utility, and under the assumptions of the purpose framework, the necessary and sufficient conditions for human-robot alignment are as follows:}

\begin{equation}
 \begin{aligned}
  &(1)\ \exists\, G_{c,i,d}^{\iota*} \subseteq f^{E-O}_{c,i}(p_{c,i}^{\iota*})\; \land \\
  &(2)\ f_{c}^{O-S}(G_{c,i,d}^{\iota*}) = S_{c,i,d}^\omega \subseteq S_{h,i,d}^*  \; \land \\
  &(3)\ f_h^{S-O}(S_{c,i,d}^\omega) = O_{h,i,d}^\omega \subseteq G_{h,i,d}^*\; \land \\
  &(4)\ f^{O-E}_{h,i}(O_{h,i,d}^\omega) = E_{h,i}^{\omega} \subseteq P_{h,i}^{\iota*}
 \end{aligned}
 \label{Eq:AppendixTheoremAlignmentExtrinsic}
\end{equation}  

\noindent\textbf{Proof.}
We first prove the sufficiency and then the necessity of the conditions for alignment as defined above.

\noindent\textbf{\textit{Sufficiency (conditions $\Rightarrow$ alignment)}}. 
To prove sufficiency, we show how the conditions guarantee a causal chain that leads to the satisfaction of the human purpose.
For the assumption on the robot intention tuple, when the human selects a purpose $P_{h,i}^{\iota*}$ and domain $d$, the robot acquires (if needed) and selects a purpose $P_{c,i}^{\iota*}$, selects a purpose point $p_{c,i}^{\iota*} \in P_{c,i}^{\iota*}$, and targets the same domain of the human $d_{c,i}^{\iota*}=d$.

Condition 1 states that the robot knows and commits to a domain-specific goal $G_{c,i,d}^{\iota*}$.
If $G_{c,i,d}^{\iota*}$ is not known, the robot must discover or learn it.
This goal completes the robot intention by specifying a concrete goal within the set of possible ones encoded by $f_{c,i}^{E-O}(p_{c,i}^{\iota*})$.  
This specification is necessary because $p_{c,i}^{\iota*}$ is an abstract purpose point, and its effective pursuit requires the selection of a domain-grounded goal $G_{c,i,d}^{\iota*}$ that is compatible with it.

Condition 2 ensures that the robot has the capability to achieve goal $G_{c,i,d}^{\iota*}$ that is enough to cause a world state $s_d \in S_{c,i,d}^\omega$ that satisfies the human desired world states, $S_{c,i,d}^\omega \subseteq S_{h,i,d}^{*}$.
This condition is important for grounding human-robot alignment in the triangular relationship that connects robot intentions and human expectations through their shared interpretation of environmental effects.

Condition 3 guarantees that every world state in $S_{c,d,i}^\omega$ is interpreted by the human, via $f_{h}^{S-O}$, as an observation in the set $O_{h,i,d}^\omega \subseteq G_{h,i,d}^{*}$.
Thus, all world states resulting from the pursuit of $G_{c,i,d}^{\iota*}$ are perceived by the human as satisfying a desirable goal.

Finally, condition 4 closes the triangulation by establishing that the human interpretation through $f_{h,i,d}^{O-E}$ of any specific outcome point $o_{h,i,d}^\omega \in O_{h,i,d}^\omega$ that the robot might cause with its action satisfies the human intended purpose $P_{h,i}^{\iota*}$.

As we have seen in the main text, together these conditions guarantee a causal-perceptual chain for which the robot's selected goal leads to world effects, which the human perceives and encodes as fulfilling the intended purpose: 

\begin{equation}
\forall P_{h,i}^{\iota*}\; \exists G_{c,i,d}^{\iota*}\left(
f_{h,i}^{O-E} \circ 
f_{h}^{S-O} \circ 
f_{c}^{O-S}(G_{c,i,d}^{\iota*})
\subseteq P_{h,i}^{\iota*}\right)
\end{equation}

\noindent\textbf{\textit{Necessity (conditions $\Leftarrow$ alignment)}}. 
We prove necessity by contradiction.  
We assume that alignment holds, while the conjunction of the conditions does not.  
By De Morgan's laws, the negation of the conjunction implies that at least one of the conditions is false.  
We thus show that if any condition is false, alignment is violated, leading to a contradiction.  
Hence, the implication `alignment $\Rightarrow$ conditions' must hold.

If condition 1 is false, the robot does not pursue a goal $G_{c,i,d}^{\iota*}$ corresponding to $p_{c,i}^{\iota*}$.
Therefore, the robot does not behave properly and the world will not satisfy the human purpose $P_{h,i}^{\iota*}$, thus violating alignment.

If condition 2 is false, either the robot pursues an incorrect goal $G_{c,i,d}^{\iota}$ or selects an aligned goal $G_{c,i,d}^{\iota*}$ but lacks sufficient competence to successfully accomplish it.  
In either case, the resulting world states $S_{c,i,d}^{\omega}$ fall outside the human-desired region $S_{h,i,d}^{*}$, and misalignment arises.

If condition 3 is false, the human interprets the world states $S_{c,i,d}^{\omega}$ possibly resulting from the robot's action as falling outside their desired goal $G_{h,i,d}^{*}$.
This might happen either because the robot has pursued a wrong goal or has not enough competence (condition 1 or 2 are false), or because the human misperceives the effects of the robot action.
In the latter case, the misalignment is due to the human rather than the robot.  

If condition 4 is false, the human interpretation of the perceived outcome in $E_{h,i}^{\omega}$ does not fall within $P_{h,i}^{\iota*}$. 
This could again be due to prior mapping failures (conditions 1-3) or to human misclassification.
In either case, the chain does not end in successful purpose recognition, and alignment is violated.

Therefore, failure of any one of the conditions 1–4 contradicts the assumption of alignment.
Hence, all conditions are necessary.

\hfill $\qedsymbol$

\subsection{\textbf{Alignment with extrinsic purpose and variable utility}}
\label{Sec:AppendixExtrinsicPurposeVariableUtilityProof}

\noindent\textbf{Definition: alignment with extrinsic purpose and variable utility.}
\textit{When the human has an extrinsic purpose $P_{h,i}^{\iota*}$ formed by points with utility values $f_{h,i}^{E-U}(p_{h,i}^{\iota*})$, human-robot alignment with respect to $P_{h,i}^{\iota*}$ and domain $d$ holds if:}
\begin{equation}
\begin{aligned}
  &\left(\mathcal{P}_{h}^{\iota*} = \{P_{h,i}^{\iota*}\} \;\land\; D_{h,i}^{\iota*} = \{d\}\right) 
  \rightarrow \\
  &\exists\, P_{c,i}^{\iota*} \;
  \exists\, p_{c,i}^{\iota*} \;
  \exists\, d_{c,i}^{\iota*}\\ 
  &\left(p_{c,i}^{\iota*} \in P_{c,i}^{\iota*} \;\land\; 
  d_{c,i}^{\iota*} = d \;\land\; f_{h,i}^{E-U}(e_{h,i}^\omega) > \theta_{h,i}^{E-U} \right)
\end{aligned}
\label{Eq:AppendixDefinitionAlignmentExtrinsicUtility}
\end{equation}
\textit{where $\theta_{h,i}^{E-U}$ is a task-dependent utility threshold above which the outcome $e_{h,i}^\omega$ is considered to satisfy human intention.}

\noindent\textbf{Theorem: alignment with extrinsic purpose and variable utility.}
\textit{In the case where the human has an extrinsic purpose $P_{h,i}^{\iota*}$ composed of points with associated utilities, and under the assumptions of the purpose framework, the necessary and sufficient conditions for human-robot alignment are:}
\begin{equation}
 \begin{aligned}
  &(1)\ \exists\, G_{c,i,d}^{\iota*} \subseteq f^{E-O}_{c,i}(p_{c,i}^{\iota*})\; \land \\
  &(2)\ f_{c}^{O-S}(G_{c,i,d}^{\iota*}) = S_{c,i,d}^\omega \subseteq S_{h,i,d}^*  \; \land \\
  &(3)\ f_h^{S-O}(S_{c,i,d}^\omega) = O_{h,i,d}^\omega \subseteq G_{h,i,d}^*\; \land \\
  &(4)\ f^{O-E}_{h,i}(O_{h,i,d}^\omega) = e_{h,i}^{\omega} \ \land\ f_{h,i}^{E-U}(e_{h,i}^{\omega}) > \theta_{h,i}^{E-U}
 \end{aligned}
\label{Eq:AppendixTheormeAlignmentExtrinsicUtility}
\end{equation}

\noindent\textbf{Proof.}
The proof follows the same structure as in the case of uniform utility purpose points (\ref{Sec:AppendixExtrinsicPurposeProof}), with the exception of condition 4.

\noindent\textbf{\textit{Sufficiency (conditions $\Rightarrow$ alignment)}}. Conditions (1)–(3) ensure that the robot selects a goal $G_{c,i,d}^{\iota*}$ compatible with the intended purpose point $p_{c,i}^{\iota*}$, pursues it competently, and causes a world state that the human perceives as a desirable goal within the purpose.
Condition 4, now stronger, ensures that the possibly perceived outcomes $e_{h,i}^{\omega}$, computed through the human mappings, exceeds the desired threshold, $f_{h,i}^{E-U}(e_{h,i}^{\omega}) > \theta_{h,i}^{E-U}$.
Thus, the robot's actions lead to outcomes that are judged by the human as sufficiently desirable.

\noindent\textbf{\textit{Necessity (conditions $\Leftarrow$ alignment)}}. If any condition fails, misalignment results. In particular, if condition 4 fails, the final outcome may either not belong to $P_{h,i}^{\iota*}$, or belong to it, but it has a too low utility to be considered acceptable by the human, hence failing alignment.
This explicitly handles cases where matching the outcome set is not sufficient and a minimum degree of satisfaction (utility) is required for alignment.

\hfill $\qedsymbol$

\subsection{\textbf{Alignment with an intrinsic purpose}}
\label{Sec:AppendixIntrinsicPurposeProof}

\noindent\textbf{Definition: alignment with intrinsic purpose}. \textit{In the case where the human has an intrinsic purpose $P_{h,i}^{\iota*}$, assumed to be composed of points with uniform utility, and a target domain $d$, human-robot alignment holds when:}

\begin{equation}
 \begin{aligned}
  &\left(\mathcal{P}_{h}^{\iota*} = \{P_{h,i}^{\iota*}\}\; \land\; D_{h,i}^{\iota*} = \{d\}\right) \rightarrow \\[5pt]
  &\forall\, G_{h,i,d}^{\epsilon} \in \{f_{h,i,d}^{E-O}(p_{h,i}^{\iota*})\}_{p_{h,i}^{\iota*} \in P_{h,i}^{\iota*}} \\
  &\left(\exists\, P_{c,i}^{\iota*}\; \exists\, p_{c,i}^{\iota*}\; \exists\, d_{c,i}^{\iota*} \left(d_{c,i}^{\iota*} = d\; \land\; 
  O_{h}^{\omega} \subseteq G_{h,i,d}^{\epsilon}\right)\right)
 \end{aligned}
 \label{Eq:DefinitionAlignmentIntrinsic}
\end{equation} 

\textit{where
$P_{h,i}^{\iota*}$ and $d$ are the desired human purpose and target domain;
$P_{c,i}^{\iota*}$ and $d_{c,i}^{\iota*}$ are the robot's intended purpose and selected domain;
$G_{h,i,d}^{\epsilon}$ is any human extrinsic goal possibly commanded by the human during the extrinsic phase;
$O_{h}^{\omega}$ is the percept of the effect of the robot action in $d$ caused by the robot pursuit of $p_{c,i}^{\iota*}$.
}

\noindent\textbf{Theorem: intrinsic purpose.}
\textit{In the case the human has an intrinsic purpose formed by points with uniform utility, and if the assumptions of the purpose framework hold, the necessary and sufficient conditions for human-robot alignment are as follows:}

\begin{equation}
 \begin{aligned}
  &(1)\ \exists\ G_{c,i,d}^\iota = f^{E-O}_{c,i,d}(p_{c,i}^{\iota*})\ \land \\
  &(2)\ f_{c}^{O-S}(G_{c,i,d}^\iota) = S_{c,i,d}^\omega \subseteq S_{h,i,d}^* \ \land \\
  &(3)\ f_{h}^{S-O}(S_{c,i,d}^\omega) = O_{h,i,d}^\omega \subseteq G_{h,i,d}^{\epsilon}
 \end{aligned}
 \label{Eq:AppendixTheoremAlignmentIntrinsic}
\end{equation}

\noindent\textbf{Proof}

We first prove the sufficiency and then the necessity of the conditions for alignment as defined above.

\noindent\textbf{\textit{Sufficiency (conditions $\Rightarrow$ alignment)}}.  
Sufficiency can be proved by showing that, for any \textit{extrinsic} goal that the human might select among those relevant to their \textit{intrinsic} purpose, the conditions imply that the robot selection of the corresponding intrinsic purpose and purpose point triggers a causal chain that leads to the achievement of that goal.

By definition of alignment, when the human selects $P_{h,i}^{\iota*}$ and $d$, these lead the robot to form (if needed), and select a corresponding purpose $P_{c,i}^{\iota*}$ and the domain $d_{c,i}^{\iota*} = d$.

Condition 1 states that the robot can discover and select a goal $G_{c,i,d}^\iota$ uniform to $f_{c,i,d}^{E-O}(p_{c,i}^{\iota*})$, that is, a goal that satisfies a purpose point of $P_{c,i}^{\iota*}$.

Condition 2 states that the robot has anough competence to pursue goal $G_{c,i,d}^\iota$ that is enough to lead to a world state in $S_{c,i,d}^\omega$, and this is within the scope of the effects $S_{h,i,d}^*$ in the world that the human desires through its intrinsic purpose point $p_{h,i}^{\iota*}$.

Condition 3 establishes that the human-perceived effects in the environment $O_{h,i,d}^\omega$, caused by the robot pursuit of $G_{c,i,d}^{\iota}$, fulfil the human goal $G_{h,i,d}^{\epsilon}$.

Together, conditions 1-3 establish \textit{a causal chain} that fulfils alignment, in particular ensuring that for each possible desired extrinsic goal $G_{h,i,d}^{\epsilon}$ that the human might desire in the future, and corresponding to a point $p_{h,i}^{\iota*}$ of the intrinsic purpose $P_{h,i}^{\iota*}$, the robot is able to select a goal $G_{c,i,d}^{\iota}$ that causes the realisation of $G_{h,i,d}^{\epsilon}$:

\begin{equation}
  \forall G_{h,i,d}^{\epsilon}\; \exists  G_{c,i,d}^{\iota}\left(f_{h}^{S-O}\circ f_{c}^{O-S}(G_{c,i,d}^{\iota})\subseteq G_{h,i,d}^{\epsilon}\right)
\end{equation}

\noindent\textbf{\textit{Necessity (conditions $\Leftarrow$ alignment)}}. 
As for the previous theorems, we prove necessity by contradiction by showing that assuming any one of the conditions as false implies the violation of the assumption of alignment, thus implying that each condition needs to be true.

Assuming condition 1 is false means that there is no $G_{c,i,d}^\iota$ that is pursued by the robot, thus the state assumed by the world will not satisfy the human intrinsic purpose $P_{h,i}^{\iota*}$ and alignment will be violated.

If condition 2 is false, the robot might pursue the goal $G_{c,i,d}^\iota$ but this leads the world to a state that does not satisfy the human purpose $P_{h,i}^{\iota*}$, either because it is a wrong goal that produces effects outside the scope of the human intrinsic purpose, or because, even if relevant, the robot's competence for it is insufficient.

Finally, assuming that condition 3 is false implies that the human interprets the state accomplished by the robot in the world as not satisfying any possible extrinsic goal $G_{c,i,d}^{\epsilon}$ relevant for the intrinsic purpose $P_{h,i}^{\iota*}$.
This could be due to the robot having selected the wrong goal $G_{c,i,d}^\iota$ or not having enough competence for it so that, although it is possibly within the scope of the human intrinsic purpose (condition 2), it does not fulfil the specific extrinsic goal $G_{h,i,d}^{\epsilon}$.
Alternatively, the misalignment might be due to human misperception/misclassification of the robot action outcome, rather than the robot behaviour.  

\hfill $\qedsymbol$

\subsection{\textbf{Alignment with an extrinsic purpose and instrumental goals}}
\label{Sec:AppendixExtrinsicPurposeInstrumentalGoalsProof}

\noindent\textbf{Definition: alignment with an extrinsic purpose and instrumental goals.}  
\textit{When the human has an extrinsic purpose composed of points with uniform utility and reachable through instrumental goals, human-robot alignment with respect to purpose $P_{h,i}^{\iota*}$ and domain $d$ holds if:}

\begin{equation}
  \begin{aligned}
  &\left(\mathcal{P}_{h}^{\iota*} = \{P_{h,i}^{\iota*}\} \;\land\; D_{h,i}^{\iota*} = \{d\}\right) 
  \rightarrow \\
  &\exists\, P_{c,i}^{\iota*} \;
  \exists\, p_{c,i}^{\iota*} \;
  \exists\, d_{c,i}^{\iota*} 
  \left(p_{c,i}^{\iota*} \in P_{c,i}^{\iota*} \;\land\; 
  d_{c,i}^{\iota*} = d \;\land\; 
  e_{h,i}^\omega \in P_{h,i}^{\iota*}\right)
  \end{aligned}
\label{Eq:AppendixDefinitionAlignmentExtrinsicInstrumental}
\end{equation}  

\textit{where
$P_{h,i}^{\iota*}$ and $d$ denote the human’s intended purpose and target domain;  
$P_{c,i}^{\iota*}$, $p_{c,i}^{\iota*}$, and $d_{c,i}^{\iota*}$ are the robot's selected purpose, purpose point, and domain of execution; 
$e_{h,i}^\omega$ denotes the human's encoding of the outcome of the robot's action, caused by pursuing $p_{c,i}^{\iota*}$.
}

\noindent\textbf{Theorem: extrinsic purpose and instrumental goals.}
\textit{In the case where the human has an extrinsic purpose composed of points with uniform utility and the robot may need to pursue it through instrumental goals, the necessary and sufficient conditions for human-robot alignment, under the assumptions of the purpose framework, are as follows:}

\begin{equation}
  \begin{aligned}
  &(1)\ \exists\, G_{c,i,d}^{\iota*} \subseteq f^{E-O}_{c,i}(p_{c,i}^{\iota*}) \;\land \\
  &(2)\ \exists\, G_{c,i,d,1}^\iota, \dots, G_{c,i,d,n}^\iota\\
  &\quad\; \left(\forall\, j = 1, \dots, n-1 \left( G_{c,i,d,j}^\iota \rightarrow \mathcal{E}(G_{c,i,d,j+1}^\iota)\right)\right)\; \land\;\\
  &\quad \;\ G_{c,i,d,n}^\iota \rightarrow \mathcal{E} (G_{c,i,d}^\iota) \;\land \\
  &(3)\ f_{c}^{O-S}(G_{c,i,d}^{\iota*}) = S_{c,i,d}^\omega \subseteq S_{h,i,d}^* \;\land \\
  &(4)\ f_h^{S-O}(S_{c,i,d}^\omega) = O_{h,i,d}^\omega \subseteq G_{h,i,d}^{*}\\
  &(5)\ f^{O-E}_{h,i}(O_{h,i,d}^\omega) = E_{h,i}^\omega \subseteq P_{h,i}^{*}
  \end{aligned}
\label{Eq:AppendixTheoremAlignmentExtrinsicInstrumental}
\end{equation}  

\noindent\textbf{Proof.}

We first prove the sufficiency and then the necessity of the conditions.
In particular, we focus only on the additional condition 2 because the rest of the demonstration is as for the extrinsic case (\ref{Sec:AppendixExtrinsicPurposeProof}).

\noindent\textbf{\textit{Sufficiency (conditions $\Rightarrow$ alignment)}}. 
Sufficiency holds if the conditions imply that the robot selection of a purpose $P_{c,i}^{\iota*}$ and a point $p_{c,i}^{\iota*}$ trigger a causal chain leading to the satisfaction of the human purpose.
Condition 2 states that, in addition to the goal $G_{c,i,d}^{\iota*}$, the robot also selects a proper sequence of goals $G_{c,i,d,1}^\iota, \dots, G_{c,i,d,n}^\iota$, each ensuring that the preconditions of the following goal are met, until $G_{c,i,d}^{\iota*}$ is achieved.
This also implies that the robot has enough competence to suitably select those goals based on the goal-selector $\Pi_c$ and to select a suitable sequence of actions to accomplish each of them through the goal-conditioned policy $\pi_c$.

\noindent\textbf{\textit{Necessity (alignment $\Rightarrow$ conditions)}}.
Necessity can be proved by contradiction by assuming that any of the conditions is false.  
In particular, if condition 2 is false, the robot fails to select suitable subgoals  $G_{c,i,d,j}^\iota$ through $\Pi_c$ or to have the necessary competence $\pi_c$ to achieve them.
Either results in the impossibility to accomplish $G_{c,i,d}^{\iota*}$, thus interrupting the causal chain leading to the satisfaction of human purpose $P_{h,i}^{\iota*}$.

\hfill $\qedsymbol$

\subsection{\textbf{Alignment with an extrinsic purpose, instrumental goals, and a proscriptive purpose}}
\label{Sec:AppendixExtrinsicPurposeInstrumentalGoalsProscriptivePurposeProof}

\noindent\textbf{Definition: alignment with an extrinsic purpose, instrumental goals, and a proscriptive purpose.}  
\textit{When the human has an extrinsic purpose composed of points with uniform utility, human-robot alignment with respect to the extrinsic purpose $P_{h,i}^{\iota*}$, the proscriptive purpose $P_{h,i}^{\xi*}$, and domain $d$ holds if:}

\begin{equation}
  \begin{aligned}
  &\left(\mathcal{P}_{h}^{\iota*} = \{P_{h,i}^{\iota*}\} \;\land\; \mathcal{P}_{h}^{\xi*} = \{P_{h,i}^{\xi*}\} \;\land\; D_{h,i}^{\iota*} = \{d\}\right) 
  \rightarrow \\
  &\exists\, P_{c,i}^{\iota*} \;
  \exists\, p_{c,i}^{\iota*} \;
  \exists\, d_{c,i}^{\iota*}\\ 
  &\Big(p_{c,i}^{\iota*} \in P_{c,i}^{\iota*} \;\land\; 
  d_{c,i}^{\iota*} = d \;\land\; 
  e_{h,i}^\omega \in P_{h,i}^{\iota*} 
  \;\land\; \\  
  &\forall j=1,\dots,n (e_{h,i,j}^\omega \notin P_{h,i}^{\xi*}) \;\land\; e_{h,i}^\omega \notin P_{h,i}^{\xi*} \Big)
  \end{aligned}
\label{Eq:AppendixDefinitionAlignmentExtrinsicIntrumentalProscriptive}
\end{equation}  

\textit{where
$P_{h,i}^{\iota*}$ and $d$ are the desired human purpose and target domain, and $P_{h,i}^{\xi*}$ a proscriptive purpose;
$P_{c,i}^{\iota*}$, $p_{c,i}^{\iota*}$ and $d_{c,i}^{\iota*}$ are the robot's pursued purpose, purpose point, and selected domain;
$e_{h,i,j}^{\omega}$ and $e_{h,i}^{\omega}$ are the human encoding of the effects of the robot actions pursuing respectively the subgoals and goal in $d$.
}

\noindent\textbf{Theorem: extrinsic purpose, instrumental goals, and proscriptive purpose.}
\textit{In the case where the human has an extrinsic purpose composed of points with uniform utility, a proscriptive purpose, and the robot may need to pursue it through instrumental goals, the necessary and sufficient conditions for human-robot alignment, under the assumptions of the purpose framework, are as follows:}

\begin{equation}
 \begin{aligned}
  &(1)\ \exists\, G_{c,i,d}^{\iota*} \subseteq f^{E-O}_{c,i}(p_{c,i}^{\iota*}) \;\land \\
  &(2)\ \exists\, G_{c,i,d,1}^\iota, \dots, G_{c,i,d,n}^\iota\\
  &\quad \;\left(\forall\, j = 1, \dots, n-1 \left( G_{c,i,d,j}^\iota \rightarrow \mathcal{E}(G_{c,i,d,j+1}^\iota)\right)\right)\; \land\;\\
  &\quad \;\ G_{c,i,d,n}^\iota \rightarrow \mathcal{E} (G_{c,i,d}^\iota) \;\land \\
  &(3)\ f_{c}^{O-S}(G_{c,i,d}^{\iota*}) = S_{c,i,d}^\omega \subseteq S_{h,i,d}^* \;\land\; \\
  &\quad \;\forall j\, (f_{c}^{O-S}(G_{c,i,d, j}^{\iota}) = S_{c,i,d,j}^\omega \nsubseteq S_{h,i,d}^{\xi*}) \;\land \\
  &(4)\ f_h^{S-O}(S_{c,i,d}^\omega) = O_{h,i,d}^\omega \subseteq G_{h,i,d}^{*}  \;\land\; \\
  &\quad \; \forall j\, (f_h^{S-O}(S_{c,i,d,j}^\omega) = O_{h,i,d,j}^\omega \nsubseteq G_{h,i,d,j}^{\xi*})\\
  &(5)\ f^{O-E}_{h,i}(O_{h,i,d}^\omega) = E_{h,i}^\omega \subseteq P_{h,i}^{*} \;\land\; \\
  &\quad \; \forall j\, (f^{O-E}_{h,i}(O_{h,i,d,j}^\omega) = E_{h,i,j}^\omega \nsubseteq P_{h,i,j}^{\xi*})
 \end{aligned} \label{Eq:TheoremAlignmentExtrinsicIntrumentalProscriptive}
\end{equation}

\noindent\textbf{Proof}.

We first prove the sufficiency and then the necessity of the conditions.
We consider only the additional second parts of conditions (2-5) because the rest of the demonstration is as for the case of extrinsic purpose and instrumental goals (\ref{Sec:AppendixExtrinsicPurposeInstrumentalGoalsProof}).

\noindent\textbf{\textit{Sufficiency (conditions $\Rightarrow$ alignment)}}. 
Sufficiency holds if the specified conditions ensure that the robot's selection of a purpose and a point within it initiates a causal chain that achieves the intended human purpose and leads the human to perceive that the forbidden purpose is being respected.
Condition 3 states that, $\forall j = 1,\dots,n$, the instrumental subgoals $_{c,i,d, j}^{\iota}$ do \textit{not} produce unwanted effects in the environment.
Condition 4 adds that the human does \textit{not} interpret these effects as observations $O_{h,i,d,j}^{\omega}$ that are undesirable.
Finally, condition 5 specifies that, at the same time, the human does \textit{not} abstract such observations into representations $E_{h,i,j}^{\omega}$ that would indicate the causation of the forbidden purpose.

\noindent\textbf{\textit{Necessity (alignment $\Rightarrow$ conditions)}}.
Necessity can be proved by contradiction by assuming that any of the conditions is false.  
If condition 3 is violated, then there exists at least one instrumental subgoal $G_{c,i,d,j}^{\iota}$ that produces an effect in the environment that is inconsistent with the human intention, thus contradicting alignment.  
If condition 4 is violated, then the human perceives the resulting environmental effect as an observation $O_{h,i,d,j}^{\omega}$ that is undesirable, thus violating human expectations.  
If condition 5 is violated, then the human abstracts these observations into a representation $E_{h,i,j}^{\omega}$ that signifies that the activation of the forbidden purpose, again contradicting alignment.  
Therefore, a violation of any of the three conditions leads to a contradiction of the alignment assumption, which proves their necessity.

\hfill $\qedsymbol$

\subsection{\textbf{Alignment with an extrinsic purpose and multiple domains.}}
\label{Sec:AppendixExtrinsicPurposeMultipleDomainsProof}

\noindent\textbf{Definition: alignment with an extrinsic purpose and multiple domains.}  
\textit{When the human has an extrinsic purpose composed of points with uniform utility, human-robot alignment with respect to such purpose $P_{h,i}^{\iota*}$ and multiple domains $D_{h,i}^{\iota*} = \{d_1, \cdots,d_j,\cdots d_n\}$ holds if:}

\begin{equation}
  \begin{aligned}
  &\left(\mathcal{P}_{h}^{\iota*} = \{P_{h,i}^{\iota*}\} \;\land\; D_{h,i}^{\iota*} = \{d_1, \cdots,d_j,\cdots d_n\}\right) 
  \rightarrow \\
  &\exists\, P_{c,i}^{\iota*} \;
  \exists\, p_{c,i}^{\iota*} \;
  \forall\, j=1,\cdots,n \\
  &\left(p_{c,i}^{\iota*} \in P_{c,i}^{\iota*} \;\land\; 
  d_{c,i,j}^{\iota*} = d_j \;\land\; 
  e_{h,i}^\omega \in P_{h,i}^{\iota*}\right)
  \end{aligned}
\label{Eq:AppendixDefinitionAlignmentExtrinsicDomains}
\end{equation}   

\textit{where  
$P_{h,i}^{\iota*}$ and $\{d_1, \cdots,d_j,\cdots d_n\}$ denote the human’s intended purpose and target domains;  
$P_{c,i}^{\iota*}$, $p_{c,i}^{\iota*}$, and $d_{c,i,j}^{\iota*}$ are the robot selected purpose, purpose point, and domains of execution;  
$e_{h,i}^\omega$ is the human's encoding of the outcome of the robot's action caused by pursuing $p_{c,i}^{\iota*}$ in any target domain.}

\noindent\textbf{Theorem: extrinsic purpose and multiple independent domains.}
\textit{In the case where the human has an extrinsic purpose composed of points with uniform utility, and under the assumptions of the purpose framework, the necessary and sufficient conditions for human-robot alignment are as follows:}

\begin{equation}
 \begin{aligned}
  \forall j=1,\dots,n
  \begin{cases}
  (1)\ \exists\, G_{c,i,d_j}^{\iota*} \subseteq f^{E-O}_{c,i}(p_{c,i}^{\iota*})\; \land \\
  (2)\ f_{c}^{O-S}(G_{c,i,d_j}^{\iota*}) = S_{c,i,d_j}^\omega \subseteq S_{h,i,d_j}^*  \; \land \\
  (3)\ f_h^{S-O}(S_{c,i,d_j}^\omega) = O_{h,i,d_j}^\omega \subseteq G_{h,i,d_j}^*\; \land \\
  (4)\ f^{O-E}_{h,i}(O_{h,i,d_j}^\omega) = E_{h,i}^{\omega} \subseteq P_{h,i}^{\iota*}\\
  \end{cases} 
 \end{aligned}
 \label{Eq:TheoremAlignmentExtrinsicDomains}
\end{equation}

\noindent\textbf{Proof.}  
The proof builds on the theorem concerning alignment with a single extrinsic purpose, and extends it by incorporating the additional requirement that alignment must hold across all human-intended target domains.  

\noindent\textbf{\textit{Sufficiency (conditions $\Rightarrow$ alignment).}}  
If the conditions hold for each domain in the set of human-intended domains, then alignment holds in each domain individually for the same reasoning illustrated in \ref{Sec:AppendixExtrinsicPurposeProof} for the single-domain case.  
Alignment holding across all target domains then ensures global alignment.

\noindent\textbf{\textit{Necessity (alignment $\Rightarrow$ conditions).}}  
The conditions are necessary because, if any of them fails in at least one of the human-intended domains, then alignment does not hold in that domain due to the same reasoning illustrated in \ref{Sec:AppendixExtrinsicPurposeProof} for the single-domain case. This implies that alignment fails overall, since the requirement is to achieve alignment in \textit{all} specified domains.

\hfill $\qedsymbol$

\bibliographystyle{elsarticle-num} 
\bibliography{Bibliography.bib}

\end{document}